\documentclass[conference]{IEEEtran}
\IEEEoverridecommandlockouts
\usepackage{cite}
\usepackage{amsmath,amssymb,amsfonts}
\usepackage{graphicx}
\usepackage{hyperref}
\hypersetup{colorlinks,allcolors=black}
\usepackage{multirow}
\usepackage{subfig}
\usepackage{float}
\usepackage{tabularray}
\usepackage{cleveref}
\usepackage{booktabs}
\usepackage{xspace}

\newcommand{\eg}{{\emph{e.g.}},\xspace}
\def\BibTeX{{\rm B\kern-.05em{\sc i\kern-.025em b}\kern-.08em
    T\kern-.1667em\lower.7ex\hbox{E}\kern-.125emX}}
\def\up{$\uparrow$}

\begin{document}

\title{Size Aware Cross-shape Scribble Supervision for Medical Image Segmentation}

\author{\IEEEauthorblockN{1\textsuperscript{st} Jing Yuan}
\IEEEauthorblockA{\textit{Department of Electrical and Electronic Engineering} \\
\textit{Imperial College London}\\
London, UK \\
j.yuan20@imperial.ac.uk}
\and
\IEEEauthorblockN{2\textsuperscript{nd} Tania Stathaki}
\IEEEauthorblockA{\textit{Department of Electrical and Electronic Engineering} \\
\textit{Imperial College London}\\
London, UK \\
t.stathaki@imperial.ac.uk}
}

\maketitle

\begin{abstract}
  Scribble supervision, a common form of weakly supervised learning, involves annotating pixels using hand-drawn curve lines, which helps reduce the cost of manual labelling.
This technique has been widely used in medical image segmentation tasks to fasten network training.
However, scribble supervision has limitations in terms of annotation consistency across samples and the availability of comprehensive groundtruth information.
Additionally, it often grapples with the challenge of accommodating varying scale targets, particularly in the context of medical images.
In this paper, we propose three novel methods to overcome these challenges, namely, 1) the cross-shape scribble annotation method; 2) the pseudo mask method based on cross shapes; and 3) the size-aware multi-branch method.
The parameter and structure design are investigated in depth.
Experimental results show that the proposed methods have achieved significant improvement in mDice scores across multiple polyp datasets.
Notably, the combination of these methods outperforms the performance of state-of-the-art scribble supervision methods designed for medical image segmentation.
\end{abstract}


\section{Introduction}
\label{sec: scribble-introduction}
In recent years, automatic medical image segmentation \cite{fan2020pranet, UNet, zhou2018unet++} has been widely used because it can be used as a diagnostic aid for doctors, thus greatly reducing their workload.
Such accurate image segmentation relies on precise large-scale annotations, which is time-consuming and painstaking.
In this case scribble supervision \cite{Lin_2016_CVPR} is proposed to accelerate labelling by annotating images with scribbles.
As there is no need to outline the silhouettes of the targets, fast scribble annotations have been investigated extensively \cite{boundary-mix, ensemble-pseudo, diffusion, zhang2023zscribbleseg}.

\begin{figure*}[tb]    
  \centering 
  \subfloat[]
  {
      \includegraphics[width=0.15\linewidth]{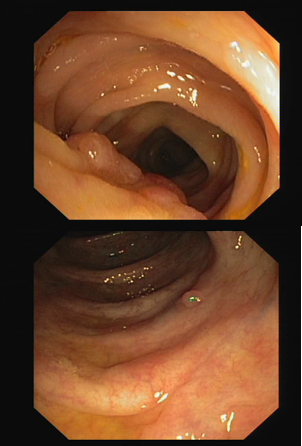}
  }
  \subfloat[]
  {
      \includegraphics[width=0.15\linewidth]{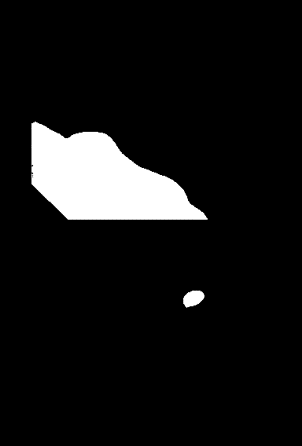}
  }
    \subfloat[]
  {
      \label{fig:subfig3}\includegraphics[width=0.15\linewidth]{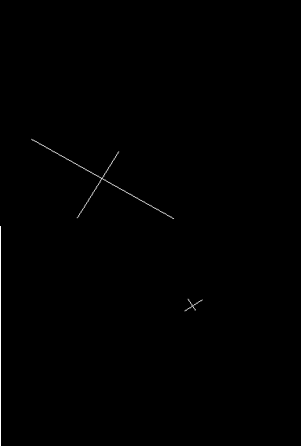}
  }
    \subfloat[]
  {
      \label{fig:subfig4}\includegraphics[width=0.15\linewidth]{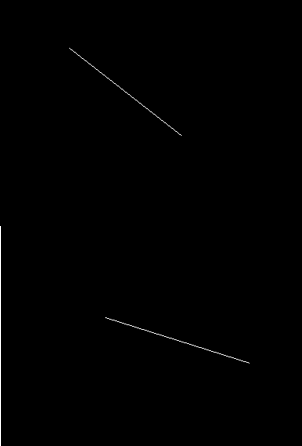}
  }
  \caption{Examples of relatively large and small polyps, the full groundtruth masks and the scribbles used in this paper to annotate the foreground and background area. (a) The images. (b) Full groundtruth masks. (c) Foreground scribble in crossing shape. (d) Background scribble.} 
  \label{fig:large and small polyps}
\end{figure*}

While scribble supervision alleviates the painful annotating process, it encounters three main challenges.
\begin{itemize}
    \item \textbf{Insufficient pixel-level annotations.}
    As only pixels on scribbles are labelled, the updating of the network may be ineffective.
    To be specific, only a few parameters affected by the labelled scribbles are optimized in each iteration.
    Additionally, the network can overfit easily.
    It may produce different results if the input image is transformed, such as re-scale and rotation.
    
    \item \textbf{Lack of consistency in annotating.}
    There is limited regularization of the scribble annotating process.
    For a single manual annotator, the relative position and appearance of the scribbles change in different targets and images.
    For multiple manual annotators, some tend to draw a circle for a round-shaped target, while some tend to draw arcs on two sides.
    Thus, it might affect the fair comparison across multiple scribble-supervision networks.
    Although the lack of regularization gives more freedom and randomness in labelling and saves the time of locating the exact pixels or even alleviates overfitting to some extent, the effectiveness of such random labelling is questioned.
    For example, scribbles might sometimes be closer to the boundary or longer for large targets, but sometimes not.
    This results in under-explored scribble information regarding relative label positions, length/size etc. considering randomness in annotating.

    \item \textbf{Unbalanced amount of labels across varying scale targets.}
    This problem widely exists in many medical images such as \Cref{fig:large and small polyps}. 
    It inherits this problem from supervised segmentation, which has long been ignored in weak supervision.
    It will lead the network to be proficient or overfit in dominant categories but ignore the minorities.
    For example, it is observed that small stuff can be easily submerged in the large background area.
\end{itemize}

Three methods are proposed in this paper to overcome these challenges.
First, we propose a novel annotating pattern by regularizing random scribbles to cross-shape segments, effectively intersecting the entire target area (\Cref{fig:large and small polyps}(c)).
Second, we generate a pseudo mask based on these cross-shape scribbles to increase the number of pixel-level annotations.
Third, the size-aware multi-branch approach is proposed to handle the segmentation of different scale targets.
This is achieved by applying three parallel branches and the proposed size aware loss which enhances the contribution according to the relative size estimated by the pseudo masks.
Although the idea of 'size-aware' has been popular in other tasks, \eg object detection, to the best of the author's knowledge, it is rarely mentioned in the field of scribble supervision.
This paper explores the application of this idea and proposes an exclusive two-stage module which demonstrates its effectiveness in scribble-supervised medical image segmentation.
Comprehensive experiments are conducted on the structure design and choice of parameters across multiple datasets.
Results show that the proposed approaches improve the performance significantly on various datasets when they are applied to different existing segmentation models.
It is observed that the proposed methods outperform state-of-the-art segmentation models which are designed for medical image segmentation with scribble supervision on multiple datasets.

Our main contributions are:
\begin{itemize}
    \item Propose a straightforward, yet innovative cross-shape scribble annotation method for medical image segmentation, which provides an effective means of regularizing scribble labels.
    \item Propose the size-aware multi-branch method to improve the segmentation accuracy of both small and large targets for the first time in the field of scribble-supervised medical image segmentation, owing to the organic integration of the novel pseudo mask generation, the size-aware loss function and the implementation of multiple branches.
    \item Investigate the design and implementation of our method in depth and achieve state-of-the-art performance on various datasets.
\end{itemize}

\section{Related Work}
Most studies try to address the problem of limited pixel-level annotations, which restricts the ability to train more effective neural networks, whether through supervised or unsupervised methods.
The main idea of them is to increase the amount of information that can be used to train a better network either in a supervised or unsupervised way.
Several approaches generate full-segmentation pseudo-labels to increase the level of supervision \cite{ensemble-pseudo, TreeLossAffinityLowHigh,regularizationLossPseudo, covid19, diffusion, reliable-label, Zhang_2020_CVPR, Vernaza_2017_CVPR, Lin_2016_CVPR}.
Others propagate the scribble information to unlabelled pixels \cite{Vernaza_2017_CVPR, diffusion, Lin_2016_CVPR}.
However, it is challenging to guarantee the accuracy of the pseudo labels as they are only learnt from scribble annotations with insufficient supervision.
Wrong pseudo-labels are ambiguous, which will confuse the network, and may result in learning wrong information.
To overcome this problem, some works combine the predicted mask with additional \cite{ensemble-pseudo, TreeLossAffinityLowHigh, covid19, diffusion, reliable-label, Zhang_2020_CVPR}.
These works usually rely on dual networks or branches to predict semantic masks in the same iteration and ensemble them following a learnt confidence map or weighted averages.
Other strategies use losses to regularize generated annotations by minimizing entropy regularization loss \cite{regularizationLossPseudo} or leveraging the additional image-level annotation \cite{scribble+image-level}, another kind of weak annotation, to enhance the supervision for the scribble-only tasks.

Apart from explicitly increasing the number of labelled pixels, some works enhance supervision by incorporating prior knowledge.
They aim to ensure that the same image, even after transformation, yields consistent prediction results.
This approach can alleviate overfitting by exploring the consistency between images to regularize predictions.
For example, \cite{shape-consistency-loss} introduces the concept of shape consistency loss, \cite{Zhang_2022_CVPR} proposes the cycle consistency, and \cite{regularizationLossPseudo} proposes the entropy regularization loss.

Some works study the implementation of specially designed data augmentation approaches.
For example, \cite{boundary-mix} substitutes another image for the predicted boundary segmentation of the input image, while \cite{zhang2023zscribbleseg} proposes supervision augmentation, which trains a network to learn efficient scribble annotations.
Additionally, \cite{Zhang_2022_CVPR} combines two images into one, which is then sent to the segmentation network along with the two original images.
Some research focuses on boundary identification and usage, as seen in \cite{boundary-mix, Zhang_2020_CVPR, wang2019boundary}.

Notably, all the aforementioned recent works effectively address the first challenge outlined in \Cref{sec: scribble-introduction}. However, the other two challenges are rarely studied in current scribble-supervision tasks.
To the best of our knowledge, only \cite{zhang2023zscribbleseg} mentions two rules of efficient scribble annotation.

\section{Method}
\subsection{Overall pipeline}
This section presents an innovative size-aware multi-branch method as shown in \Cref{fig:saf architecture}.
This method consists of two modules, namely the Pseudo Mask Generation (PMG) module and the Two-stage Multiple Branches (TMB) module in which the size-aware loss function is implemented.
In the proposed method, the input image is first fed into the PMG module to generate the corresponding pseudo masks, then sent to the backbone to extract semantic features.
These masks and features are both sent to the TMB module for two-stage processing.
Finally, the optimal predicted mask will be selected from the results of multiple branches produced in the first (train segmentation) stage according to the learned confidence scores through the second (train score) stage. 

\subsection{Pseudo Mask Generation (PMG)}
\begin{figure*}[tb]   
  \centering 
  \subfloat[]
  {
      \includegraphics[width=0.24\linewidth]{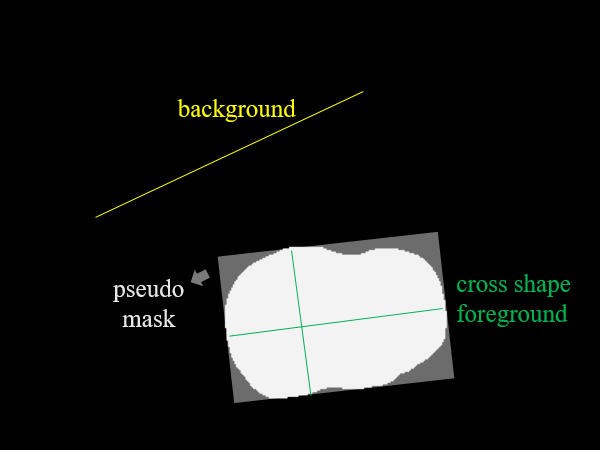}
  }
  \subfloat[]
  {
      \includegraphics[width=0.6\linewidth]{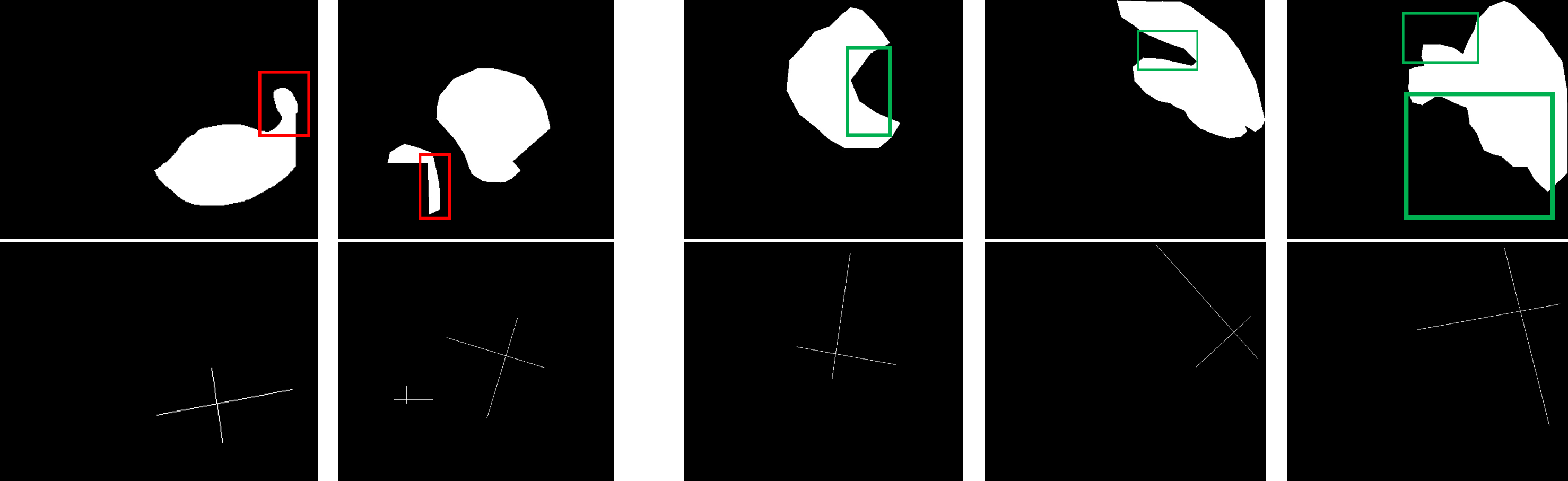}
  }
  \caption{(a) The example of the cross-shape scribble and the corresponding pseudo mask of a polyp.
   The foreground(polyp) is annotated with two nearly perpendicular line segments (green).
   The background is annotated with a single line segment (yellow).
   All the scribbles have a width of one pixel.
   The generated pseudo mask (grey) is the outer parallelogram of the cross-shape scribble.
   (b) Examples of cross-shape scribble annotations for twisted polyps used in this paper.
   The upper row shows the full masks.
   The bottom row shows the cross-shape scribble annotations.
   The red and green boxes mark the areas that will be ignored or mistakenly included after applying the pseudo mask generated by the rough cross-shape scribbles.}   
  \label{fig:cross shape scribbling and pmask}
\end{figure*}

PMG takes two steps to produce a pseudo mask.
First, efficient cross-shape scribbling is proposed to annotate the targets.
Second, based on such annotations, the pseudo masks are automatically generated following a pre-designed mask function.
\paragraph{Cross-shape scribbling} The proposed scribbling method employs two nearly perpendicular straight line segments, namely the cross shape, to annotate the foreground target, while a single line segment is used to mark the background region.
Examples of these scribbles are shown in \Cref{fig:large and small polyps} and \Cref{fig:cross shape scribbling and pmask}.
The segments do not require accurate alignment with the 'width' and 'length' of the target, but they only need to roughly mark the main part of the target.
In this paper, twisted polyps are either only partially covered, \eg the first and second columns of \Cref{fig:cross shape scribbling and pmask}(b), or some negative areas are also covered, \eg the third to fifth columns of \Cref{fig:cross shape scribbling and pmask}(b). 
The influence of the roughness of labelling is discussed in \Cref{sec: experiments}.
As only straight lines are allowed, cross-shape labelling is easier and faster than hand-crafted curves which usually imitate the silhouettes of the target. 
The annotator needs only select the two endpoints and the rest of the segment can be automatically filled by labelling software, saving the effort to painstakingly draw the entire curve.

\begin{figure}[tb]
  \centering
   \includegraphics[width=0.6\linewidth]{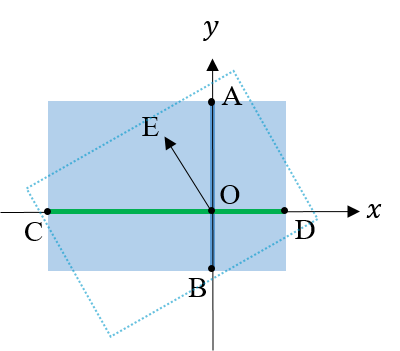}
   \caption{The sketch of the generation of the pseudo mask based on cross-shape scribbles. A weighted mask is initialized within the region defined by the crossing scribbles as depicted by the blue shaded area. Subsequently, this mask is rotated to align with the orientation of the target, as shown in the dotted area.
   }
   \label{fig:gen pseudo mask}
\end{figure}

\paragraph{Generation of the pseudo mask}
\label{sec:pseudo mask}
As shown in \Cref{fig:gen pseudo mask}, suppose the cross-shape scribbles are segments $AB$ and $CD$ which intersect at point $O$. 
Consider the direction of vector $\overrightarrow{OA}$ as the direction of the initial mask, denoted as $M'$. 
Suppose the real direction of the target, for example, the polyp, is determined by the unit vector $\overrightarrow{OE}$.
The final pseudo mask $M$ is generated by rotating the initial mask to the real direction of the target.
The main content presents the generation of the initial mask by multiplication, and other ways to generate the initial mask are discussed in the supplementary material. 
The resulted initial mask $\boldsymbol{M'_{\times}}$ is calculated as
\begin{equation}
    \boldsymbol{M'_{\times}}\left( x,y\right) =\exp \left[ -\left( \dfrac{x^{2}}{\sigma _{i}^{2}}+\dfrac{y^{2}}{\sigma _{j}^{2}}\right) \right].
    \label{eq:x}
\end{equation}
where $\sigma_{i{j}} (i,j=1$ or $2)$ are pre-designed parameters, and $x, y$ are positions on the mask obeying
\begin{equation}
    x\in \begin{cases}  [0,\left | OD \right | ]& \text{} i=1 \\  [-\left | OC \right |,0 ]& \text{} i=2\end{cases} ;y\in \begin{cases}  [0,\left | OA \right | ]& \text{} j=1 \\  [-\left | OB \right |,0 ]& \text{} j=2\end{cases},
\end{equation}
where $|\cdot|$ stands for the length of the segment.
Then the initial mask is rotated to the direction of $\overrightarrow{OE}$ using the corresponding rotation matrix.

\paragraph{Advantage}
In scribble supervision, the loss function is generally chosen to be partial (binary) cross entropy (hereinafter referred to as the partial loss), \eg only the loss values of the labelled pixels are calculated.
This slows down training and introduces substantial uncertainty due to the lack of reliable groundtruth information.
This problem is alleviated with our novel pseudo mask.
Utilizing such masks on the groundtruth produces substantial and relatively accurate information from the beginning of training.
This approach mitigates the errors that can arise from the progressive generation of pseudo masks, a technique often used in prior research, where scribble ground truth is expanded using previously predicted masks.
Even though there are uncovered or falsely covered pixels in the pseudo mask, they make up a relatively small portion and can be distinguished with the guidance of the dominating correct ground truth.

\subsection{Two-stage Multiple Branches (TMB)}
\label{sec: size aware multi-branch method}
\begin{figure*}[tb]
  \centering
   \includegraphics[width=1.0\linewidth]{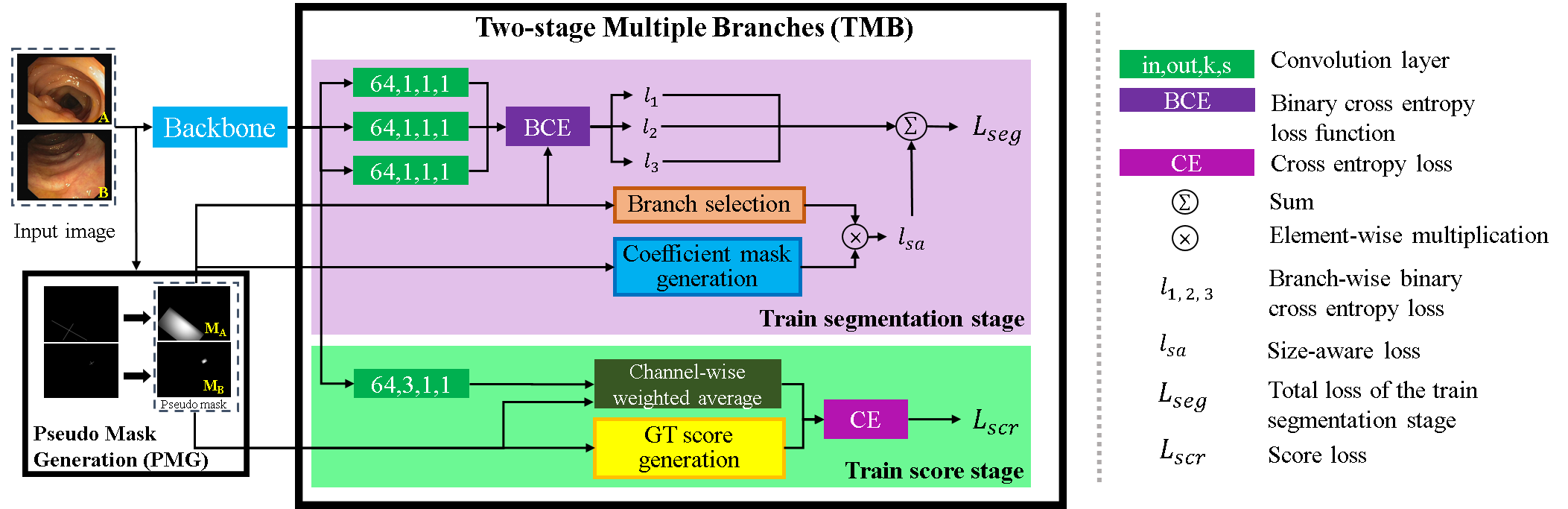}

   \caption{The architecture of the proposed size-aware multi-branch method.
   The network consists of the Pseudo Mask Generation (PMG) module and the Two-stage Multiple Branches (TMB) module.
   The TMB module comprises the train segmentation stage and the train score stage.
   The train segmentation component generates accurate prediction results through multiple branches.
   Subsequently, the backbone and the train segmentation component are held constant with updates applied exclusively to the parameters of the train score component.
   This updated component predicts confidence scores, indicating which branch is best suited to produce the optimal mask. 
   The detailed structure of the Coefficient Mask Generation (CMG), Branch Selection (BS), GT Score Generation (GSG) and the Channel-wise Weighted Average (CWA) blocks are presented in \Cref{sec: size aware multi-branch method}.
   }
   \label{fig:saf architecture}
\end{figure*}

This method consists of the train segmentation stage and train score stage.
The architecture design of this method is presented in \Cref{fig:saf architecture}.
To illustrate the process, let's consider polyp segmentation as an example.
In this scenario, the input images ($A, B$), which contain two categories, namely the poly and the background, are fed into the backbone, such as the UNet.
\paragraph{Train segmentation stage}
In the train segmentation stage, our approach is geared towards predicting precise masks while incorporating a size-aware loss function within a novel multi-branch framework.
More specifically, the backbone is followed by three convolutional layers.
From up to down these branches are indexed as 1, 2 and 3. 
They are responsible for predicting the segmentation masks ($\boldsymbol{P_{1,2,3}}$) of small, medium and large targets respectively.
These masks and pseudo masks $M_{A, B}$ are used to calculate the binary cross entropy loss ($l_{1,2,3}$) for each branch.
To achieve and strengthen the ability of prediction of varying size targets, the pseudo masks are fed into the Branch Selection (BS) and Coefficient Mask Generation (CMG) modules.
In the BS module, the relative size $r_{s}$ is computed as the ratio $r_{s}=N_{p}/N$ where $N_{p}$ is the number of positive pixels and $N$ is the number of all pixels.
The index of the selected branch is determined by
\begin{equation}
  index=\left\{\begin{matrix} 1 & r_{z}<=thr_{1} \\ 2 & thr_{1}<r_{z}<=thr_{2}\\ 3 & else\end{matrix}\right.,
\end{equation}
where $thr_1, thr_2$ are pre-designed thresholds.
They are as follows: First, calculate the $r_z$ of all training images in one or several epochs. 
Then, sort the $r_z$ value and choose threshold values, $thr_1, thr_2$ that can equally divide the non-zero $r_z$ into three segments.
In the case of the polyp training set, $thr_1=0.078, thr_2=0.177$ are utilized.
With the index of the selected branch, we obtain the corresponding BCE loss without any reduction denoted as $\boldsymbol{L'}$.
For example, if branch 1 is selected, then the BS module produces the none-reduction version of $l_1$, which is responsible for small targets.
\paragraph{Size-aware loss function}
In the CMG module, the coefficient value $\alpha$ is calculated by $\alpha =min\left (1/r_{z}, coe \right)$ where $coe$ is the pre-defined upper limit which is investigated in \Cref{sec: experiments}.
The coefficient mask $\boldsymbol{M^c}$ is generated by
\begin{equation}
    M^c_{i,j}=\left\{\begin{matrix} 0 & M_{i,j}=0\\  \alpha -1 & M_{i,j}>0\end{matrix}\right.,
\end{equation}
where $M^c_{i,j}$ and $M_{i,j}$ are the values of the coefficient mask and the pseudo mask, such as $\boldsymbol{M'_{\times}}$, at $i$th row and $j$th column respectively.
Then the size-aware loss $l_{sa}$ is expressed as
\begin{equation}
    l_{sa}=sum(\boldsymbol{L'} \cdot \boldsymbol{M^c})/N.
\end{equation}
The total loss for the train segmentation stage is
\begin{equation}
\label{eq: Lseg}
    L_{seg} = l_{sa} + l_1+ l_2+ l_3.
\end{equation}

\paragraph{Train score stage}
To select the best-predicted mask from the multiple branches, we create the train score stage to learn the confidence scores which serve as indicators.
In this stage, the backbone and the three convolution layers in the train segmentation stage remain fixed.
Only the score loss is utilized to fine-tune the parameters. 
More precisely, a single convolution layer is applied to predict the score map $\boldsymbol{S'}$ for the three branches.
The score map comprises a single score vector with three scores at each pixel.
Each score indicates the likelihood of each branch producing the best result, \eg a score vector of $(0.1,0.6,0.3)$ implies that the second branch is the optimal choice.
The Channel-wise Weighted Average (CWA) module produces the weighed average score vector $\boldsymbol{s}\in \mathbb{R}^{3}$, also referred to as the confidence scores in this paper, by
\begin{equation}
    s_i = sum(\boldsymbol{S'_i} \cdot \boldsymbol{M}) / N_p,
\end{equation}
where $i=1,2,3$, $s_i \in \boldsymbol{s}$, $\boldsymbol{S'_i}$ is the $i$th channel of the score map $\boldsymbol{S'}$, $\boldsymbol{M}$ denotes the pseudo mask, $\cdot$ is element-wise multiplication, and $N_p$ is the number of positive pixels in the pseudo mask $\boldsymbol{M}$.
The weighted average score vector stands for the communal choice of all pixels in the semantic feature map.
The GT Score Generation (GSG) manually generates the groundtruth score vector $\boldsymbol{s^g}\in \mathbb{R}^{3}$ which indicates the branch with the lowest BCE loss among the loss value of all the three branches. The $i$th ($i=1,2,3$) element of $\boldsymbol{s^g}$ is calculated as
\begin{equation}
    s^g_{i}=\left\{\begin{matrix} 1 & l_i=min(l_j|j=1,2,3)\\  0 & otherwise\end{matrix}\right..
\end{equation}
The score loss $L_{scr}$ is the cross entropy between $\boldsymbol{s^g}$ and $\boldsymbol{s}$.
In the inference stage, the branch with the highest score in the weighted average score vector will be chosen to produce the final predicted segmentation mask.

\section{Experiments}
\subsection{Setup}
\paragraph{Dataset and evaluation metric}
We conduct experiments on three widely used datasets, namely Kvasir \cite{jha2020kvasir}, CVC-ClinicDB \cite{bernal2015clinicdb} and the ACDC dataset \cite{acdc}.
The first two are polyp datasets, the last one is a general medical image dataset.
The \textbf{CVC-ClinicDB dataset} is from 29 different colonoscopy sequences, while the \textbf{Kvasir dataset} is sourced from the initial Kvasir dataset \cite{pogorelov2017kvasir}, which is asserted to be suitable for general segmentation purposes.
Kvasir contains 1,000 polyp images and CVC-ClinicDB contains 612 images from 31 colonoscopy clips.
Following \cite{fan2020pranet}, the training set is a combination of 900 images from the Kvasir dataset and 550 images from the CVC-ClinicDB dataset.
This combined training set is used for training unless otherwise specified.
The Kvasir test set contains 100 images and the CVC-ClinicDB test set contains 62 images.
Only 0.903\% and 0.11\% of pixels in the foreground and background of the combined training set are annotated with the proposed scribbling strategy. 
More details of the scribble annotations are presented in the supplementary material.

The \textbf{ACDC dataset} comprises a total of 150 examinations conducted on different patients.
Each examination consists of multiple 2-dimensional cine-MRI images obtained with two MRI scanners under various magnetic strengths and resolutions \cite{Zhang_2022_CVPR}.
It is a general medical image dataset which has been widely used to evaluate scribble supervision methods \cite{Zhang_2022_CVPR, shape-consistency-loss, ensemble-pseudo}.
The training set and the testset have 1902 and 1076 images, respectively.
The ACDC dataset has four categories (including the background).
The third category is entirely enclosed within the second category, making the latter in a ring-like shape.
Details of the scribble annotations are presented in the supplementary material.
In this paper, we randomly choose 15\% of the images in the training sets as the validation set \cite{fan2020pranet}.
For metric, the mean Dice (denoted as mDice) score is utilized to evaluate the model performance for polyp segmentation \cite{fan2020pranet}.
mDice score is between 0 and 1, with a higher mDice representing better performance.

\paragraph{Implementation details}
Our model is implemented with Pytorch Toolbox \cite{pytorch} and trained on a single NVIDIA GeoForce RTX 3090 GPU with a batch size of 4.
Adam optimizer along with the cosine annealing schedule are applied.
For Kvasir and CVC-ClinicDB datasets, the learning rates for the two stages are set as 3e-5 and 5e-4 respectively.
For the ACDC dataset, the learning rates for the two stages are both set as 1e-4.
For fair comparisons, both training and testing images are resized to $352\times352$ as previous polyp segmentation methods.
Data augmentation techniques including random rotation in the range of $[-10^{\circ},10^{\circ}]$ and color jitter with brightness, contrast and saturation all set as 0.4 are applied.
The input RGB images are normalized with means of $[0.485,0.456,0.406]$ and standard deviations of $[0.229, 0.224, 0.225]$.
UNet \cite{ronneberger2015unet} is used as the backbone if not specified.
During the inference stage of the size-aware multi-branch approach, the branch indicated by the highest weighted average (confidence) score is chosen to produce the final prediction mask.

\subsection{Ablation study}
\label{sec: experiments}
\paragraph{Effectiveness of PMG and TMB modules}
As shown in \Cref{table: ablation PMG TMB}, with the implementation of the pseudo mask generation module, the mDice scores on both test sets are increased substantially compared to the sole use of the cross-shape scribbles.
The use of the two-stage multiple branches module further improves the mDice scores on both datasets compared to the sole application of the PMG module.
Notably, the mDice scores of the proposed modules for the Kvasir test set even surpass the result achieved under the training with the full ground truth mask.
Samples of segmentation results are shown in the supplementary material.

\begin{table}
\centering
\caption{Ablation studies on the proposed Pseudo Mask Generation (PMG) module and the Two-stage Multiple Branches (TMB) module.}
\label{table: ablation PMG TMB}
\begin{tabular}{ccccc} 
\toprule
\multirow{2}{*}{Annotation} & \multicolumn{2}{c}{Module} & \multirow{2}{*}{CVC-ClinicDB} & \multirow{2}{*}{Kvasir}  \\ 
\cmidrule{2-3}
                            & PMG        & TMB           &                               &                          \\ 
\midrule
full                        & \textbackslash{}  &\textbackslash{}  & 0.82                          & 0.699                    \\
scribble                    & \textbackslash{} & \textbackslash{} & 0.5105                        & 0.523                    \\
scribble                    & \checkmark &\textbackslash{} & 0.755 (0.2445\up)             & 0.72~(0.197\up)          \\
scribble                    & \checkmark & \checkmark    & 0.771~(0.2605\up)             & 0.733~(0.21\up)          \\
\bottomrule
\end{tabular}
\end{table}

\paragraph{Influence of the roughness of labelling}
\Cref{table: shrink rate} presents the influence of the roughness of the pseudo mask on the performance.
The roughness is simulated by shrinking the lengths of the two intersecting segments by the shrink rates $r_s$ of 0.1, 0.3 and 0.5, which equals reducing the mask area to 81\%,49\% and 25\% w.r.t. the original pseudo mask.
Results from heavily shrunk masks were less satisfactory, but it's important to note that this was for a specific click supervision task and not our paper's focus.
Compared to the mDcie score of using the plain scribbles in \Cref{table: ablation PMG TMB}, the PMG module brings improvements as long as roughly 49\% or more of the target area are covered by the pseudo mask.
Utilizing relatively rough masks and scribbles still yielded competitive results, offering more flexibility in annotation.

\begin{table}[tb]
\centering
\caption{mDice scores obtained by PM method under different shrink rates ($r_s$).}
\begin{tabular}{@{}clccc@{}}
\toprule
$r_s$      & 0    & 0.1  & 0.3  & 0.5  \\ \midrule
CVC-ClinicDB & 0.76 & 0.72 & 0.54 & 0.28 \\
Kvasir   & 0.72 & 0.66 & 0.51 & 0.32 \\ \bottomrule
\end{tabular}
\label{table: shrink rate}
\end{table}

\paragraph{GT score generation}
In the training score stage, the generation of groundtruth scores can be accomplished in two distinct manners.
One is to take the branch with the lowest segmentation loss as the correct choice, as described in \Cref{sec: size aware multi-branch method}.
This straightforward approach opts for the predicted mask most likely to yield the highest mDice score.
The other is to select the branch that matches the size of the target as the correct choice.
The results are compared in the first and second rows of \Cref{table: samb-ablation}, and it appears that choosing the branch with the lowest loss leads to superior performance.

\paragraph{Channel-wise weighted average}
In the Channel-wise Weighted Average (CWA) module, the pseudo mask serves as the weighted mask to implement element-wise multiplication with the predicted scores.
To investigate the effectiveness of this mask, \Cref{table: samb-ablation} first and fourth rows compare the performance with and without the use of the pseudo mask.
It should be mentioned that if the pseudo mask is applied in the CWA module, it should also be multiplied by the predicted score map before averaging in the inference stage. Without this step, the score map does not require a weighted average.
The results clearly demonstrate that overall performance on both test sets improves when guided by the pseudo mask.

\paragraph{Additional loss}
In \Cref{table: samb-ablation}, the first and third rows compare the use of the BCE loss and the Dice loss in calculating $l_1,l_2$ and $l_3$ in the train segmentation stage.
The use of BCE loss leads to a more significant improvement in performance.
\paragraph{Structure of score prediction}
\Cref{table: samb-ablation} first and fifth rows compare two network structures applied to predict the scores in the train score stage.
The first structure is a single convolution layer.
The second structure appends a batch normalization layer, a ReLU activation function and an additional convolution layer after the first structure.
This structure has more parameters and flexibility.
Results show that a more complicated structure leads to similar mDice scores on both test sets.
This observation might be attributed to the relatively straightforward relationship between the scores and masks, making the complex structure prone to overfitting and limiting its flexibility. 
Therefore, to conserve computational resources, it is recommended to opt for the single convolution layer.

\begin{table*}[tb]
\centering
\caption{Ablation studies on the design of the TMB module.
'max', 'match': Take the branch with the least loss or with the matched relative size as the groundtruth in the GSG module.
'Dice', 'BCE': Apply dice loss or bce loss in the train segmentation stage.
'pos', 'all': Apply a pseudo mask or no mask to the weighted average in the CWA module.
'C', 'CBRC': Structures to predict the score map in train score stage.}
\label{table: samb-ablation}
\begin{tabular}{c|cc|cc|cc|cc} 
\toprule
structure          & max~             & match            & Dice             & BCE              & pos              & all              & CVC-ClinicDB & Kvasir  \\ 
\midrule
\multirow{4}{*}{C} & \checkmark       & \textbackslash{} & \textbackslash{} & \checkmark       & \checkmark       & \textbackslash{} & 0.7710       & 0.7331  \\
                   & \textbackslash{} & \checkmark       & \textbackslash{} & \checkmark       & \checkmark       & \textbackslash{} & 0.7410       & 0.7305  \\
                   & \checkmark       & \textbackslash{} & \checkmark       & \textbackslash{} & \checkmark       & \textbackslash{} & 0.7556       & 0.7270  \\
                   & \checkmark       & \textbackslash{} & \textbackslash{} & \checkmark       & \textbackslash{} & \checkmark       & 0.7744       & 0.7181  \\ 
\cmidrule(lr){1-9}
CBRC               & \checkmark       & \textbackslash{} & \textbackslash{} & \checkmark       & \checkmark       & \textbackslash{} & 0.7708       & 0.7327  \\
\bottomrule
\end{tabular}
\end{table*}

\paragraph{Choice of $coe$}
\Cref{table: coe} presents the impact of the upper limit $coe$ of the coefficient value.
Results show that $coe=10$ achieves the best mDice scores on both test sets, which greatly exceeds the case of $coe=1$ when no size awareness is introduced.
This demonstrates the effectiveness of the application of the idea of size awareness in the field of scribble-supervised medical image segmentation.

\begin{table}
\centering
\caption{Influence of the upper limit of the coefficient value $coe$ in coefficient mask generation module.}
\label{table: coe}
\begin{tabular}{ccccc} 
\toprule
$coe$        & 1       & 5      & 10     & 15      \\ 
\midrule
CVC-ClinicDB & 0.75608 & 0.7540 & 0.7710 & 0.7657  \\
Kvasir       & 0.72624 & 0.7185 & 0.7331 & 0.7210  \\
\bottomrule
\end{tabular}
\end{table}

\paragraph{Number of branches}
\Cref{table: number of branches} compares the mDice scores obtained with different numbers of branches in the train segmentation stage of the TMB module.
When only a single branch is used, this approach is the same as the pseudo mask method.
It is observed that the case of three branches achieves the overall best mDice scores on both test sets.
Reducing the number of branches to one results in the loss of the network's size-aware capability.
When there are too many branches, for example, four, the training set may not be sufficient enough to support effective learning of all branches.
Moreover, more computational resources and longer inference time are required.

\begin{table}
\centering
\caption{Influence of the number of branches.'Num.' The number of branches in size aware multi-branch approach.}
\label{table: number of branches}
\begin{tabular}{ccccc} 
\toprule
Num.         & 1     & 2      & 3      & 4       \\ 
\midrule
CVC-ClinicDB & 0.755 & 0.7723 & 0.7710 & 0.7414  \\
Kvasir       & 0.720 & 0.7051 & 0.7331 & 0.7215  \\
\bottomrule
\end{tabular}
\end{table}

\paragraph{Comparison with sota}
\Cref{table: sota} compares the state-of-the-art methods to the proposed methods combined with commonly used UNet \cite{UNet} and UNet++ \cite{zhou2018unet++} structures.
The Dual-branch \cite{ensemble-pseudo}, ShapePU \cite{shape-consistency-loss} and SCNet \cite{zhou2023SC-Net} methods are specifically designed for weakly supervised segmentation of medical images.
Results show that the proposed method can be effectively applied to different existing methods and improve performance significantly.
Moreover, the proposed method outperforms the sota methods on both CVC-ClinicDB and Kvasir test sets.
\begin{table}[tb]
\centering
\caption{Comparison with sota methods on CVC-ClinicDB and Kvasir test sets.
\textsuperscript{*}: Input image in training downsampled by 1.3 to save computational memory.
The best mDice scores of both test sets are in \textbf{bold}.
All methods use UNet as the backbone.}
\label{table: sota}
\begin{tabular}{ccc} 
\toprule
Method                              & CVC-ClinicDB   & Kvasir           \\ 
\cmidrule(lr){1-3}
UNet \cite{UNet}                    & 0.5105         & 0.523            \\
UNet++\textsuperscript{*}\cite{zhou2018unet++}  & 0.4718         & 0.6071           \\
Dual-branch \cite{ensemble-pseudo}       & 0.582          & 0.579            \\
SCNet \cite{zhou2023SC-Net}           & 0.464          & 0.477            \\
ShapePU \cite{shape-consistency-loss}          & 0.653          & 0.701            \\
UNet+proposed                       & \textbf{0.771} & 0.733            \\
UNet++\textsuperscript{*} +proposed & 0.6863         & \textbf{0.7844}  \\
\bottomrule
\end{tabular}
\end{table}

\paragraph{Generalization on ACDC dataset}
\Cref{table: supp acdc} compares the mDice scores obtained with the proposed method with the baseline UNet method on three categories of the ACDC dataset. 
It is observed that both modules improve the performance of all the categories significantly.
The idea of size awareness is highly effective as the accuracy for all the categories increases dramatically compared to the sole implementation of PMG module.
This underscores the robustness and versatility of our proposed method, which extend beyond the confines of simple binary-category polyp datasets.
Overall, the proposed method can also be applied to the segmentation of multiple categories, even with inter-class occlusions among individual pseudo masks.
The details of the design of the PMG and TMB modules for the general ACDC dataset as well as samples of segmentation results are described in the supplementary material.
\begin{table}
\centering
\caption{Effectiveness of the proposed PMG and TMB modules on the three categories (Cat1, Cat2 and Cat3) of ACDC testset.
The best mDice scores are in \textbf{bold}.}
\label{table: supp acdc}
\begin{tabular}{cc|ccc} 
\toprule
PMG              & TMB              & Cat1                                    & Cat2                                     & Cat3                                      \\ 
\cmidrule(lr){1-5}
\textbackslash{} & \textbackslash{} & 0.382                                   & 0.351                                    & 0.632                                     \\
\checkmark       & \textbackslash{} & 0.500 (0.118\up)         & 0.502 (0.15\up)           & 0.730 (0.098\up)           \\
\checkmark       & \checkmark       & \textbf{0.525(0.143\up)} & \textbf{0.539 (0.188\up)} & \textbf{0.763 (0.131\up)}  \\
\bottomrule
\end{tabular}
\end{table}

\section{Discussion}
Compared to commonly used curved scribbles, cross-shaped lines offer several distinct advantages, such as 1) they are more regularized, reducing the effect of tricks in annotation; 2) they achieve fast and precise pseudo mask generation, taking a mere 2 minutes compared to the 24 hours required by SCNet, thus significantly enhancing efficiency; 3) they produce additional area size information inspiring further exploration.
Although improvements have been achieved owing to these advantages, the proposed method has three potential problems.
First, the predicted boundaries are observed to be not identical to the groundtruth.
This might be the result of error accumulation introduced during the training using the rigid and rough rectangular pseudo masks.
This could be mitigated by incorporating more target-shape information, such as the silhouettes.
Second, it is observed that extremely small or large targets may be missed, which might be partially due to the clipped coefficient value.
Third, the labelling of extremely irregular shapes using cross-shaped lines may introduce more errors and uncertainty. 
Although the proposed method is effective on the ACDC dataset, which contains ring-like patterns, we consider more challenging and general patterns for completeness. 
In extremely odd-shaped images of organs, such as multi-gap and multi-twist patterns, the subsequent rectangular masks might either mistakenly cover some negative pixels or miss some positive pixels.
To adapt to these scenarios, two approaches might be effective: 1) apply hybrid scribbling of both curve and cross-shape lines; 2) generate multi-shape pseudo masks, \eg round, triangle, other than the sole use of rectangle ones.
Examples of failed cases are presented in the supplementary material.

\section{Conclusion}
In this paper, we propose a novel cross-shape scribble annotation strategy as a form of regularization.
Leveraging this strategy, the accurate pseudo mask and the size of the target can be estimated, leading to the pseudo mask and size-aware multi-branch approaches.
These approaches augment the availability of valuable information, consequently enhancing the segmentation performance for targets of varying scales.
Experiments show that the proposed methods when applied to different segmentation models, achieve stable and significant improvements on various polyp datasets.
Moreover, the proposed methods outperform the state-of-the-art scribble supervision methods on tested datasets.
As a future direction, we aim to extend the adaptability of our approach to general object segmentation, which often involves irregular shapes and diverse outlines.

\clearpage
\appendix
\section{Pseudo Mask Generation (PMG)}
\subsection{Two other ways for pseudo mask generation}
While the main content generates the initial pseudo mask using multiplication operation, in this section, the addition and maximization operations are presented. 
The resulted initial masks $M'_+$ and $M'_{max}$ are calculated as
\begin{equation}
    M'_{+}\left( x,y\right) =\left[ \exp \left( -\dfrac{x^{2}}{\sigma _{i}^{2}}\right) +\exp \left( -\dfrac{y^{2}}{\sigma _{j}^{2}}\right) \right] /2,
    \label{eq:+}
\end{equation}
and
\begin{equation}
    M'_{\max }\left( x,y\right) =\max \left( \exp \left( -\dfrac{x^{2}}{\sigma _{i}^{2}}\right) ,\exp \left( -\dfrac{y^{2}}{\sigma _{j}^{2}}\right) \right),
    \label{eq:max}
\end{equation}
where $\sigma_{i{j}} (i,j=1$ or $2)$ are pre-designed parameters, and $x, y$ are positions on the mask obeying
\begin{equation}
    x\in \begin{cases}  [0,\left | OD \right | ]& \text{} i=1 \\  [-\left | OC \right |,0 ]& \text{} i=2\end{cases} ;y\in \begin{cases}  [0,\left | OA \right | ]& \text{} j=1 \\  [-\left | OB \right |,0 ]& \text{} j=2\end{cases},
\end{equation}
where $|\cdot|$ stands for the length of the segment.

\begin{figure}[htb]
  \centering
   \includegraphics[width=0.8\linewidth]{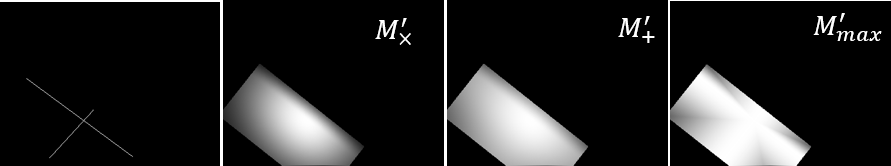}
   \caption{The cross-shape scribble and the resulting pseudo masks generated by multiplication, addition, and maximization respectively.
   }
   \label{fig:types of pseudo mask}
\end{figure}
\Cref{fig:types of pseudo mask} compares the final pseudo masks generated with multiplication, addition and maximization operations.

\subsection{Influence of the pseudo mask generation ways}
In this paper, three methods for generating the pseudo masks are presented.
To investigate the influence of these generation strategies on performance, we compare their mDice scores with the same set of $\sigma$s. \Cref{table: pseudo mask-types} shows that 'maximization' produces the best result on the CVC-ClinicDB dataset, but its performance on two test sets is unbalanced.
'Addition' produces the same mDice score as 'maximization' on the Kvasir test set, but its performance on the first dataset is far worse.
Compared to 'addition', 'multiplication' produces slightly overall better mDice scores with the most robust results on these two test sets.
Consequently, the next part investigates the 'maximization' and 'multiplication' in depth.
\begin{table}[tb]
\caption{Comparison of performance of different generation approaches of the pseudo masks.
'$\times$', '+' and 'max' represent the 'multiplication', 'addition' and 'maximization' methods respectively.
The ratios of $\sigma_{i(j)}$ are set as $|OD|,|OC|$ ($|OA|,|OB|$).}
\centering
\begin{tabular}{@{}ccc@{}}
\toprule
Type & CVC-ClinicDB & Kvasir \\ \midrule
$\times$    & 0.706        & 0.713  \\
+    & 0.711        & 0.697  \\
max  & 0.779        & 0.697  \\ \bottomrule
\end{tabular}

\label{table: pseudo mask-types}
\end{table}

\subsection{Influence of \texorpdfstring{$\sigma$}{sigma} in PMG module}
The influence of the choice of $\sigma$s on the performance of different pseudo mask generation operations is outlined in \Cref{table: pseudo mask-sig to r max and mul}.

For maximization operation, the tested $\sigma$ led to high mDice scores (close to or well above 0.7).
With the increase of $\sigma$, the difference of the mDice score on two test sets increases dramatically and then decreases in terms of the overall trend.
Interestingly, when $\sigma$ is infinity, this difference becomes slightly larger than for smaller $\sigma$ (for example when $\sigma/r$=0.75 or 2.0), but it remains within an acceptable range.
Results also indicate that with relatively high robustness, larger $\sigma$ achieves higher mDice scores.

For multiplication operation, it is observed that the increase of $\sigma$ enhances the performance and also the robustness.
Although the difference between the mDice scores on the two datasets is slightly larger when $\sigma$ reaches infinity, it remains within the tolerable range.
Notably, the mDice scores achieved on both test sets are the highest in this case.

\begin{table}
\centering
\caption{Influence of the ratio $\sigma/r$ to the performance of pseudo mask with maximization (max) and multiplication ($\times$) generation approaches.
$\sigma$ refers to $\sigma_{i(j)}$ in a general sense, $r$ refers to the corresponding length of $|OA|,|OB|,|OC|,|OD|$.
'inf' represents infinity, indicating that the pseudo mask becomes an even mask with a value of 1.0. 
}
\label{table: pseudo mask-sig to r max and mul}
\begin{tabular}{ccccc} 
\toprule
\multirow{2}{*}{$\sigma/r$} & \multicolumn{2}{c}{max}    & \multicolumn{2}{c}{$\times$}   \\ 
\cmidrule(lr){2-5}
                            & CVC-ClinicDB     & Kvasir           & CVC-ClinicDB     & Kvasir            \\ 
\midrule
0.5                         & \textbackslash{} & \textbackslash{} & 0.5449           & 0.480             \\
0.75                        & 0.715            & 0.718            & \textbackslash{} & \textbackslash{}  \\
1.0                         & 0.779            & 0.697            & 0.7062           & 0.713             \\
1.5                         & 0.740            & 0.691            & 0.720            & 0.667             \\
2.0                         & 0.747            & 0.731            & 0.714            & 0.713             \\
inf                         & 0.755            & 0.72             & 0.755           & 0.720             \\
\bottomrule
\end{tabular}
\end{table}

\begin{figure}[tb]

  \centering
   \includegraphics[width=\linewidth]{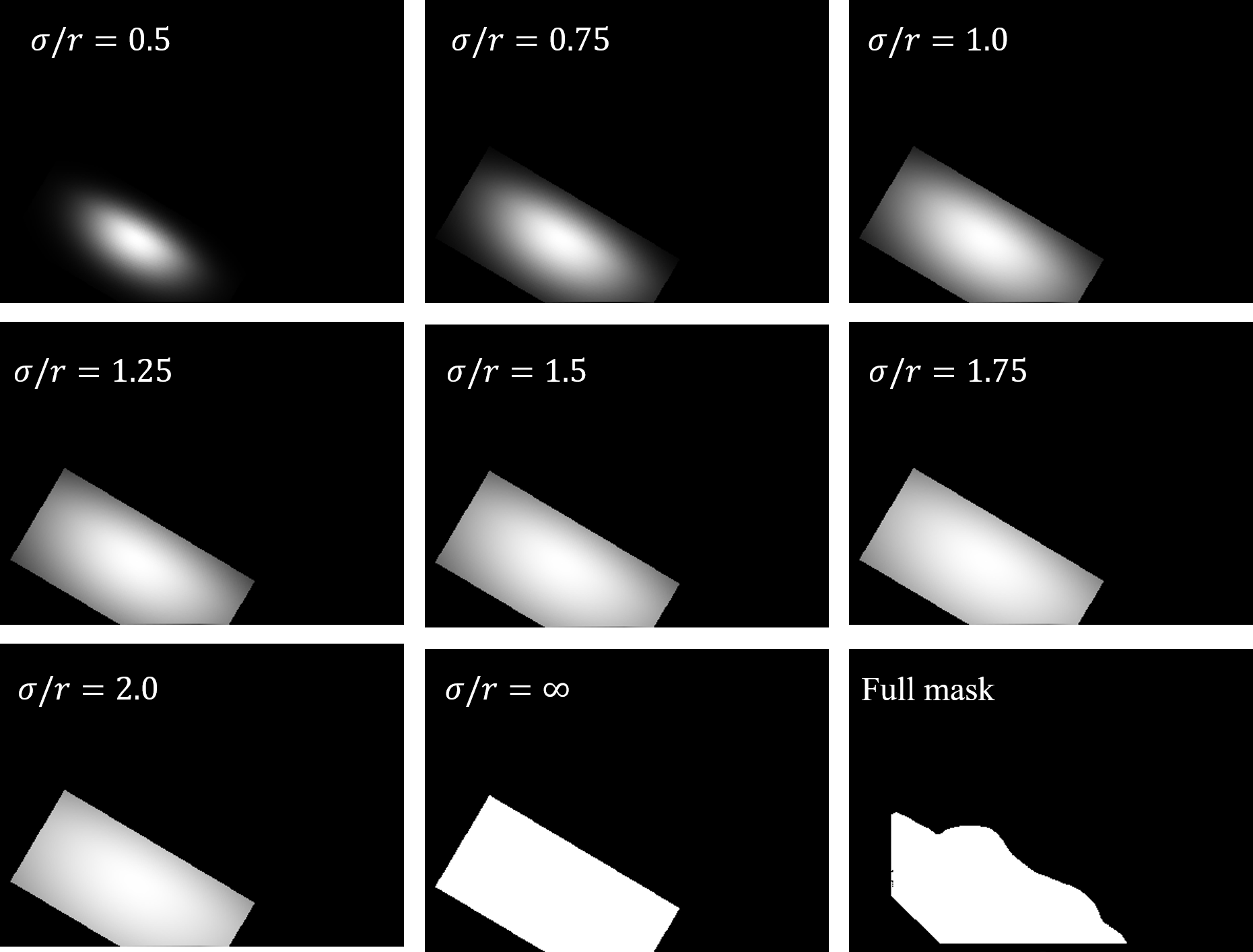}
      \caption{The first eight images are pseudo masks with different ratios $\sigma/r$.
   These masks are generated with multiplication operation.
   The last mask is the groundtruth full mask of the target.
   }
   \label{fig:xpmask sigs}
\end{figure}

To gain deeper insights into the influence of $\sigma$ on the final performance, it's essential to consider that the best performance would be achieved if the pseudo mask precisely matched the ground truth full mask. 
Therefore, studying the relative error between these two masks can provide valuable insights into this influence.
\Cref{fig:xpmask sigs} visually compares the full mask, the groundtruh for supervised learning, and pseudo masks generated with different choices of $\sigma$.
\Cref{table: relative error} presents the relative errors of the positive ($e_p$) and negative ($e_n$) pixels under different $\sigma/r$ ratios.
The errors $e_p$ and $e_n$ are calculated as $e_p=sum((M^f-M)\cdot M)/sum(M)$ and $e_n=sum((M^f-M)\cdot (I-M))/sum(I-M)$ where $M^f$ is the groundtruth full mask with value 1 at positive pixels and 0 at negative pixels, $I$ is the mask with value 1 at all pixels.
It's important to note that the presented errors are the average values across all training images.
\Cref{table: relative error} presents that the increase of $\sigma$ reduces the foreground error of the foreground dramatically from 0.76 to 0.043, which benefits the learning of foreground characteristics.
For negative error, though it is increased as more pixels are assigned as foreground, the largest $e_n$ is only 0.044 which is very small.
In addition, the relative errors of the foreground and background pixels are approximately equal only when $\sigma$ approaches infinity.
This indicates that the two categories (foreground and background) have balanced and effective information for learning.

To reach the balance of robustness and high accuracy, the $\sigma$ is chosen as infinity in subsequent experiments unless explicitly specified otherwise.
In this case, 'multiplication', 'addition' and 'maximization' operations produce the same pseudo mask which has the value 1 at all positive pixels and 0 at negative ones. 

\begin{table}[tb]
\caption{The relative error of positive ($e_p$) and negative ($e_n$) pixels between the pseudo and groundtruth masks of the polyp training set.
Pseudo masks are produced by multiplication with different $\sigma/r$ ratios.
Pseudo-masks with smaller errors are closer to the ground truth and more accurate.}
\centering
\begin{tabular}{@{}ccc@{}}
\toprule
$\sigma/r$ & $e_p$ & $e_n$ \\ \midrule
0.5   & 0.763    & 0.001  \\
0.75  & 0.555    & 0.006   \\
1.0   & 0.400    & 0.014   \\
1.25  & 0.298    & 0.021   \\
1.5   & 0.232    & 0.026   \\
1.75  & 0.188    & 0.030   \\
2.0   & 0.157    & 0.033    \\
inf   & 0.043   & 0.044    \\ \bottomrule
\end{tabular}%
\label{table: relative error}
\end{table}

\section{Evaluation on Polyp dataset}
\subsection{Details of the Cross-shape scribble annotated Polyp dataset}
\label{sec: supp dataset}
\begin{figure*}[ht]
  \centering            
  \subfloat[] 
  {
      \includegraphics[width=0.13\linewidth]{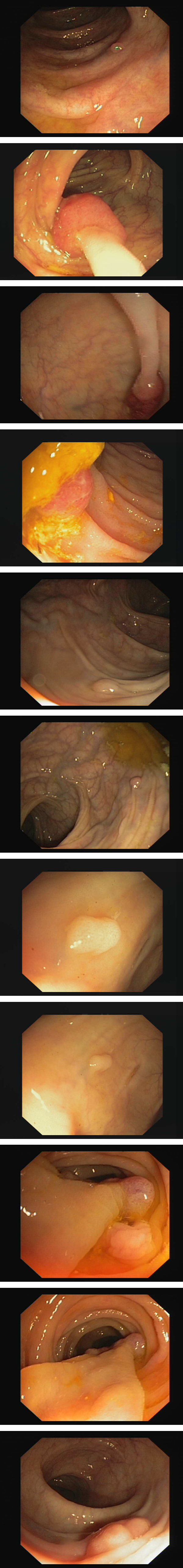}
  }
  \subfloat[]
  {
      \includegraphics[width=0.13\linewidth]{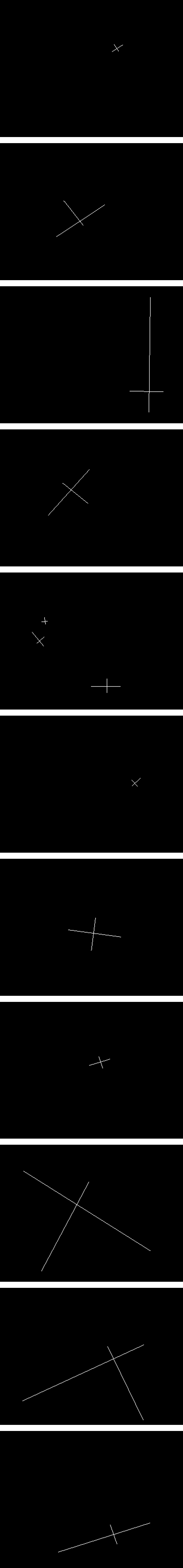}
  }
    \subfloat[]
  {
      \includegraphics[width=0.13\linewidth]{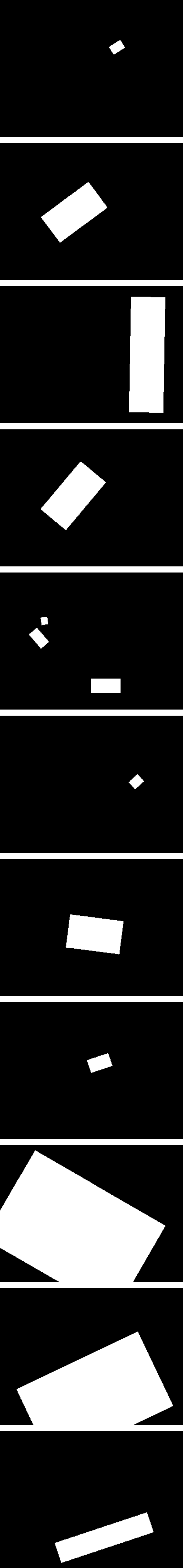}
  }
    \subfloat[]
  {
      \includegraphics[width=0.13\linewidth]{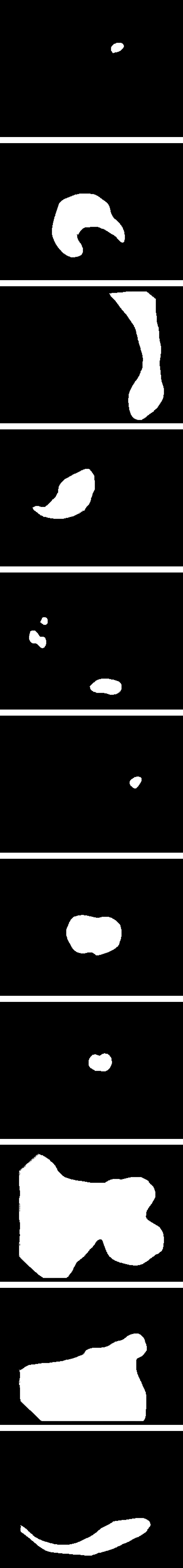}
  }
    \subfloat[]
  {
      \includegraphics[width=0.13\linewidth]{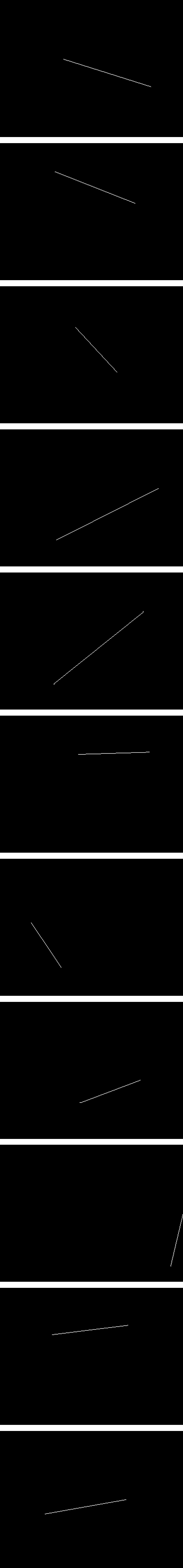}
  }
  \caption{Examples cross-shape annotated polyp images. (a) Images. (b) Foreground scribbles in crossing shape. (c) Pseudo masks. (d) Full groundtruth masks. (e) Background scribbles.}
  \label{fig: supp examples of annotated images}          
\end{figure*}

We utilize Paint Tool SAI 2, a free software that allows the annotator to edit images easily and draw segments quickly.
The process is straightforward and involves two simple steps: firstly, select one endpoint of the desired segment, and secondly, press and hold the shift key while selecting the location of the other endpoint.
Thus, the straight line with two manually selected endpoints is automatically created.
The width of the segments is one pixel.
The foreground, for example, each polyp, is marked with two segments, and the background is marked with a single segment.
Some annotated images are shown in \Cref{fig: supp examples of annotated images}.

\begin{figure*}[ht] 
  \centering            
  \subfloat[]   
  {
      \includegraphics[width=0.3\linewidth]{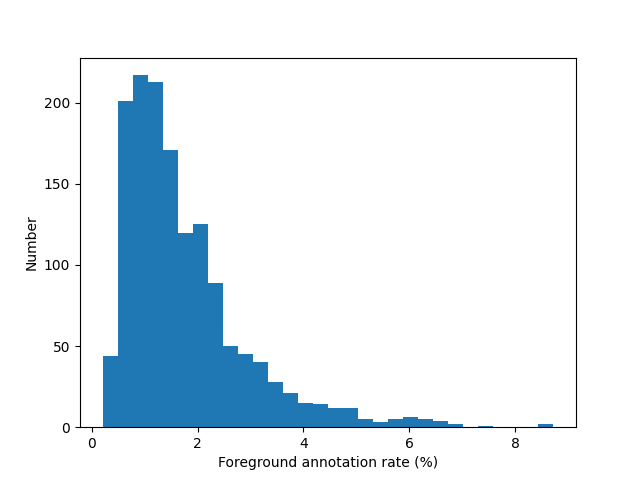}
  }
  \subfloat[]
  {
      \includegraphics[width=0.3\linewidth]{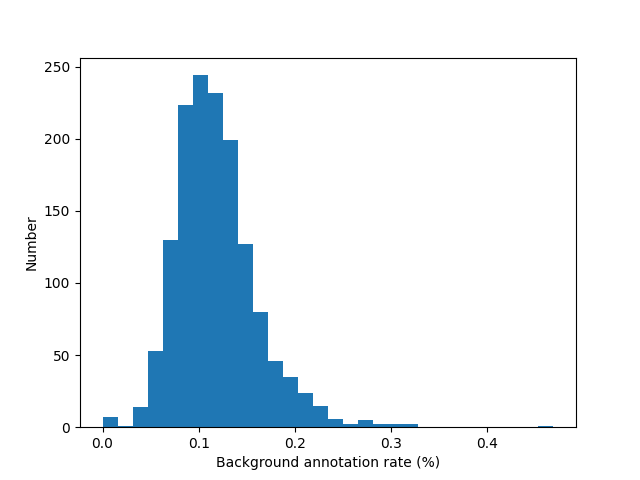}
  }
    \subfloat[]
  {
      \includegraphics[width=0.3\linewidth]{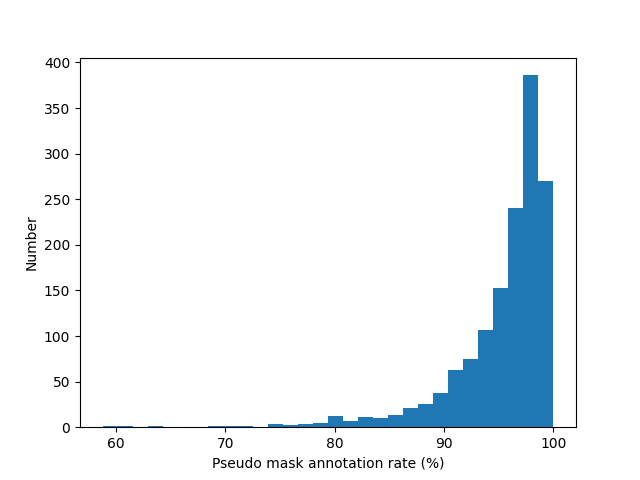}
  }
  \caption{Foreground (a) and background (b) annotation rate distributions with cross-shape annotation. (c) Foreground annotation rate distribution with generated pseudo masks.}
  \label{fig: supp distrubution}          
\end{figure*}

The combined polyp dataset used in this paper contains 1450 training images.
0.903\% and 0.11\% of pixels in the foreground and background are annotated with the proposed scribbling strategy.
\Cref{fig: supp distrubution} shows the distribution of the annotated rate of each image. 
The average annotated rate is 1.75\% and 0.12\% for foreground and background areas, respectively.
The annotated rate is calculated as the ratio of the number of labelled pixels to that of all pixels within a certain category.
The average annotated rate is the mean of annotated rates across all training images.
With the implementation of the pseudo-mask approach, 95.99\% of foreground pixels are correctly annotated.
\Cref{fig: supp distrubution} (c) shows the distribution of the annotated rate of each image.
The average annotated rate across the dataset is 95.37\%.

\subsection{Examples of results from Polyp datasets}
\label{sec: supp samples}
\begin{figure*}[ht]
    \centering
  \subfloat[]   
  {
      \includegraphics[width=0.5\linewidth]{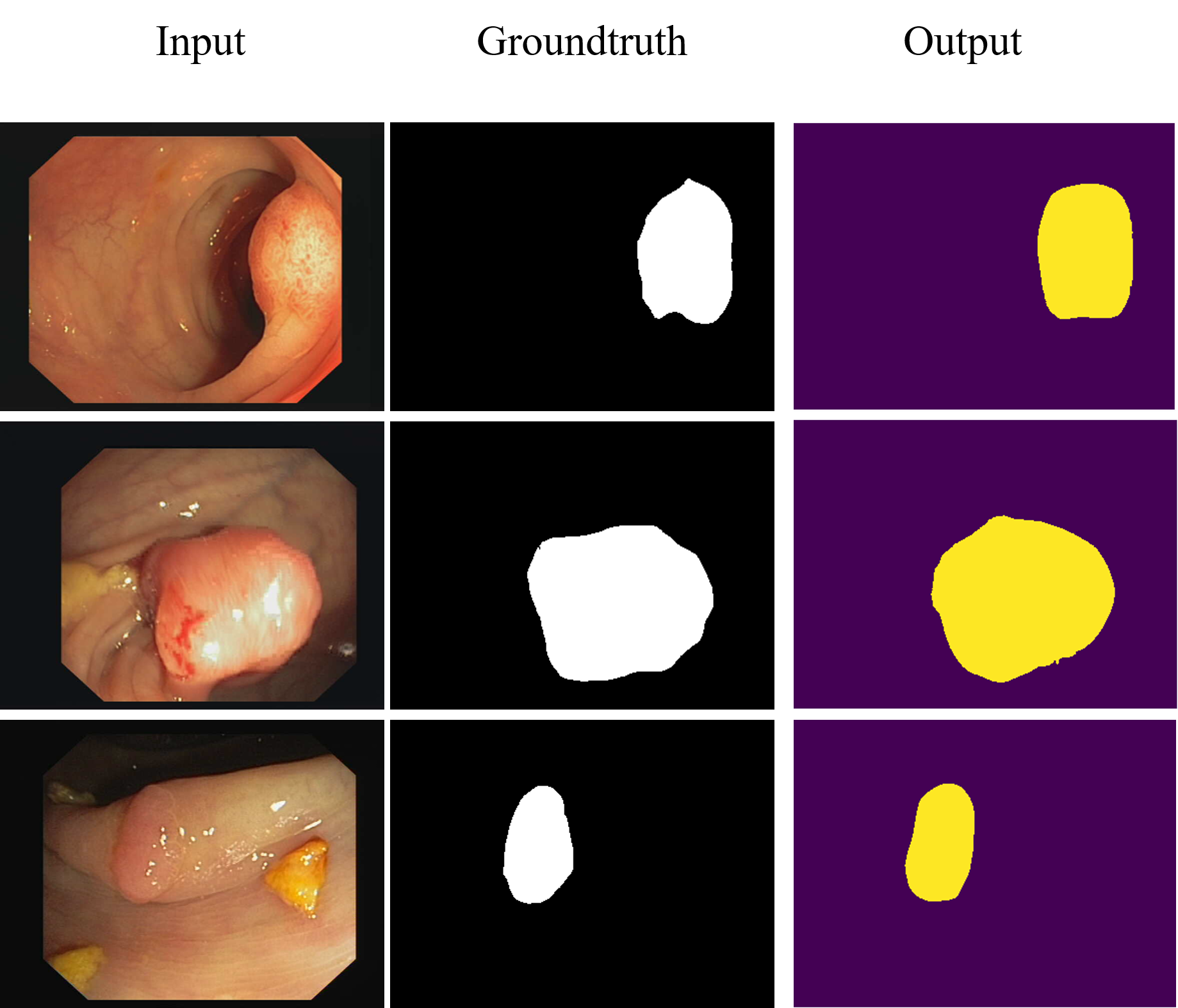}
  }
  \subfloat[]
  {
      \includegraphics[width=0.4\linewidth]{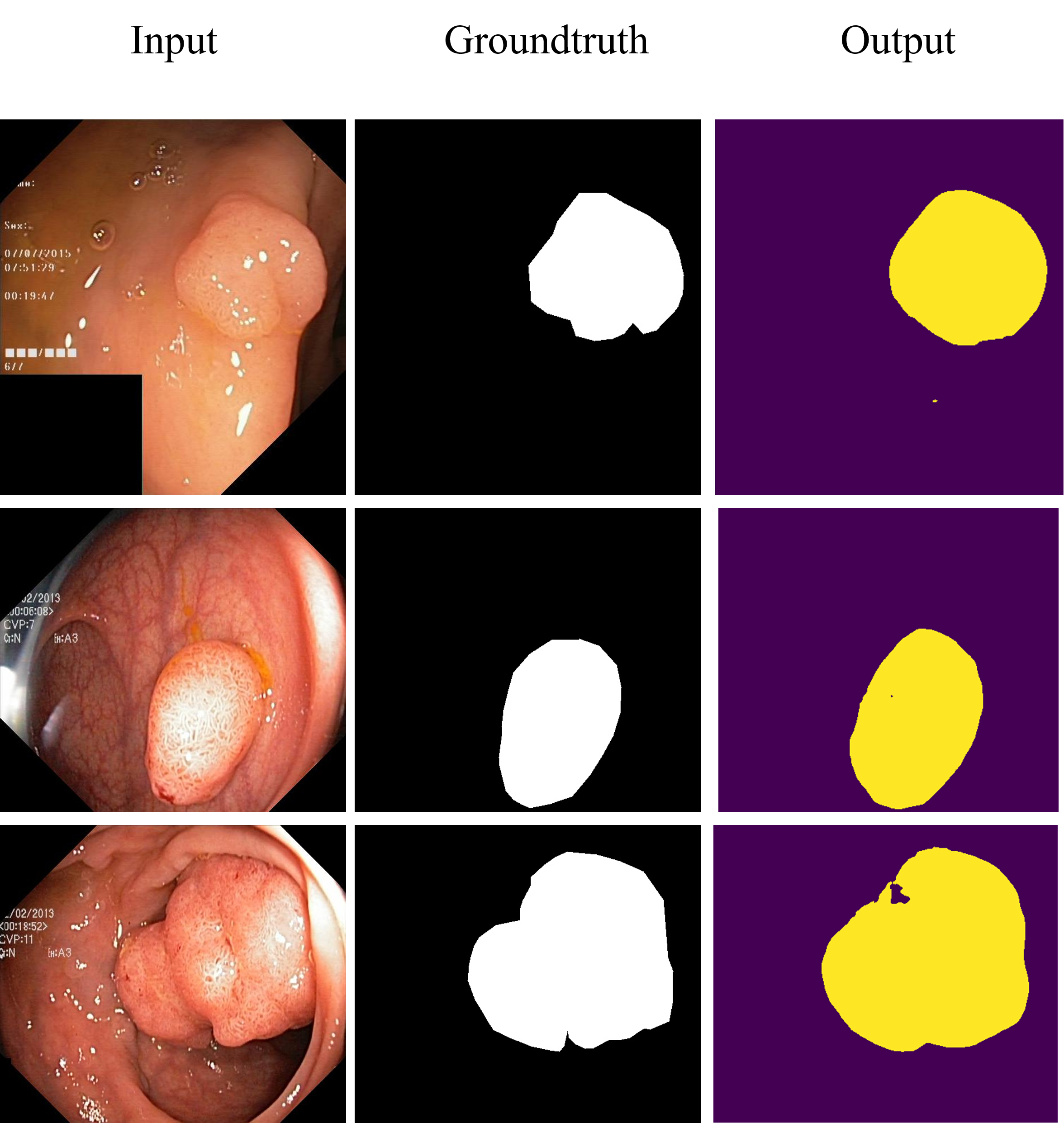}
  }
  \caption{Examples of outputs from CVC-ClinicDB test set (a) and Kvasir test set (b).}
  \label{fig: supp success samples}

\end{figure*}

\begin{figure*}
    \centering
  \subfloat[]   
  {
      \includegraphics[width=0.4\linewidth]{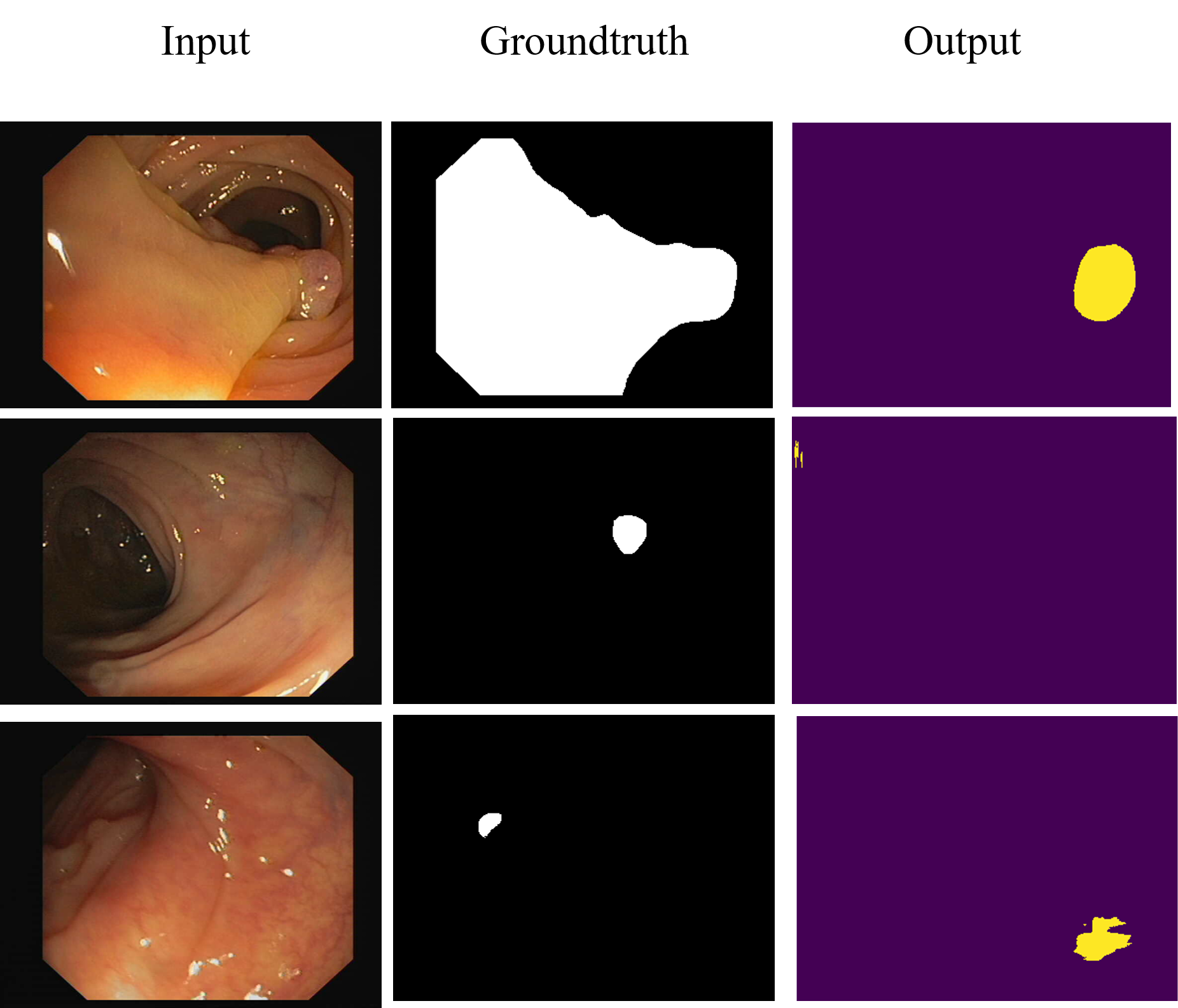}
  }
  \subfloat[]
  {
      \includegraphics[width=0.5\linewidth]{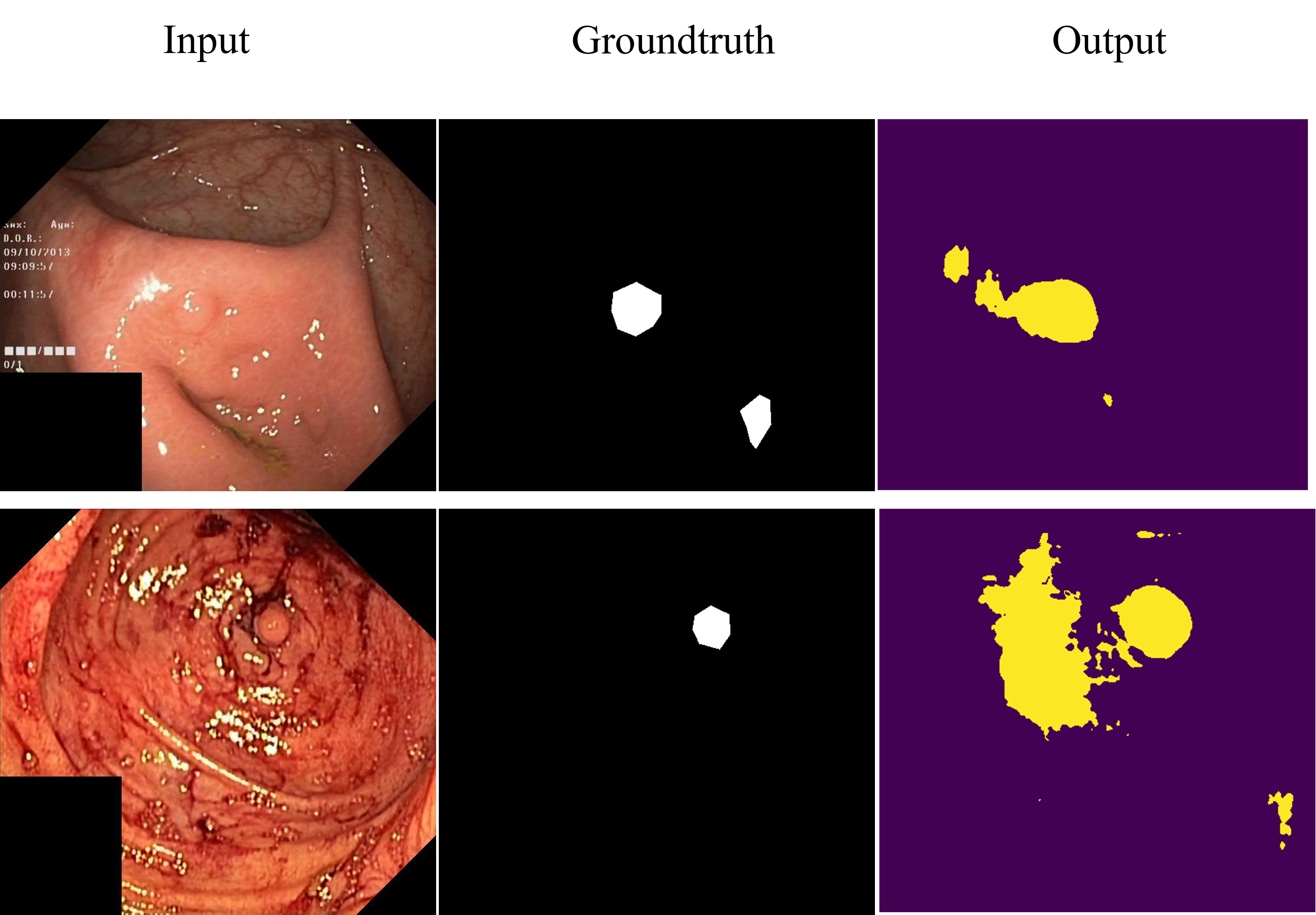}
  }
  \caption{Examples of failed outputs on CVC-ClinicDB testset (a) and Kvasir testset (b).}
  \label{fig: supp failed samples}
\end{figure*}

\begin{figure*}[ht]
    \centering
    \includegraphics[width=1.0\linewidth]{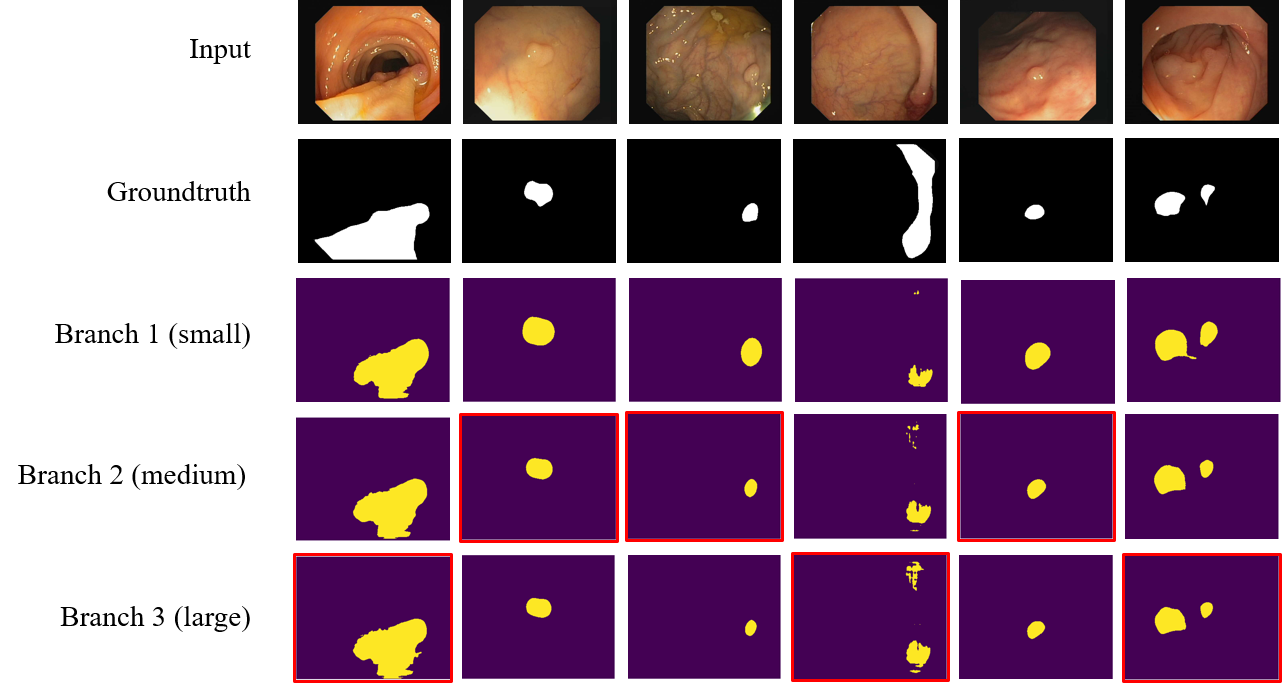}
    \caption{Examples of branch-specific outputs on CVC-ClinicDB testset. The output of each input image with the highest mDice score among the three branches is marked with a red bounding box. }
    \label{fig:supp branch-wise CVC outputs}
\end{figure*}

\Cref{fig: supp success samples} presents some successful segmentation results with the proposed method.
In case of cherry-pick, \Cref{fig: supp failed samples} shows some failed samples on two datasets.
Although the proposed size-aware multi-branch method enhances the mDice score to some extent, the segmentation of extremely large or small targets is still challenging.
To further understand how each branch facilitates the final output, \Cref{fig:supp branch-wise CVC outputs} shows the output from each branch.
The branch with the highest mDice score is bounded in red.
It is observed that for relatively large targets, the third branch, responsible for segmenting large targets, produces the best output.
The second branch tends to provide the best result for relatively small or medium-sized targets in these examples.

\section{Generalization on ACDC dataset}
\subsection{Introduction of ACDC dataset}
\begin{figure*}[htbp]
    \centering
    \includegraphics[width=0.6\linewidth]{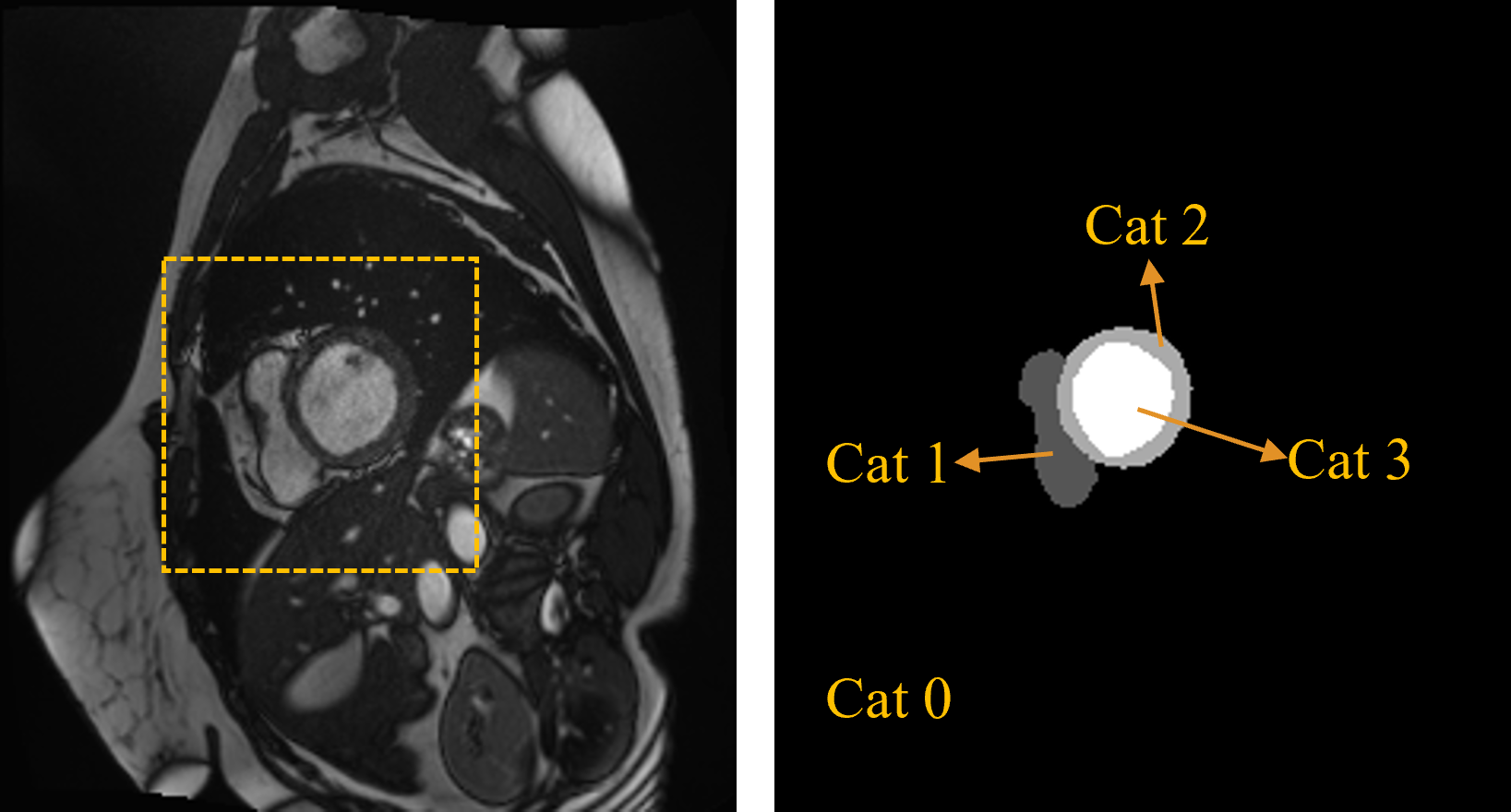}
    \caption{A typical sample of images from the ACDC dataset and its corresponding groundtruth mask.
    \textbf{Left:} One input image. The yellow bounding box shows the approximate region of the targets.
    \textbf{Right:} The groundtruth full mask. Cat 0 denotes the background area, while Cat 1, 2, and 3 denote the three categories requiring segmentation.
    }
    \label{fig: supp acdc sample}
\end{figure*}

The ACDC dataset is derived from real clinical examinations, encompassing a diverse range of well-defined pathologies and enough cases to facilitate effective training of machine learning methods. 
The dataset comprises a total of 150 examinations conducted on different patients.
Each examination consists of multiple 2-dimensional cine-MRI images obtained with two MRI scanners under various magnetic strengths and resolutions \cite{Zhang_2020_CVPR}.
It is a general medical image dataset which has been widely used to evaluate scribble supervision methods \cite{Zhang_2022_CVPR, shape-consistency-loss, ensemble-pseudo}.
This open dataset is available at \href{https://www.creatis.insa-lyon.fr/Challenge/acdc/databases.html}{this website}.
The ACDC dataset has been divided into the training set with 100 patients' examinations, while the remaining 50 examinations form the testset.
The training set and the testset have 1902 and 1076 images, respectively.
The ACDC dataset has four categories, including the background, as presented in \Cref{fig: supp acdc sample}.
Among them, 'Cat 0' denotes the background area. The other three categories, Cat 1, Cat 2 and Cat 3, require pixel-level segmentation.
As a general medical image dataset, the ACDC is more challenging than the Polyp datasets.
This is mainly due to three reasons.
First, instead of dealing with a single 'polyp' category, we are tasked with the classification of three distinct categories.
Second, the parts of these categories are connected, which can be regarded as a strong inter-class occlusion.
For example, the part of Cat 1 is connected with the part of Cat 2 while Cat 3 is entirely enclosed within Cat 2.
Third, due to occlusion, the shapes of some parts are abnormal, which requires a more delicate strategy for cross-shape scribble annotation and the subsequent pseudo mask generation process.
To adapt to the demands of general medical image segmentation, we revisit the methods for scribble annotation, network design and loss function design in the forthcoming sections.
Finally, experimental results using the ACDC testset are presented to demonstrate the effectiveness of the proposed approaches for general medical image segmentation tasks. 

\subsection{Cross-shape scribble annotated ACDC dataset}
\begin{figure*}[htbp]
  \centering
  \subfloat[]   
  {
      \includegraphics[width=0.089\linewidth]{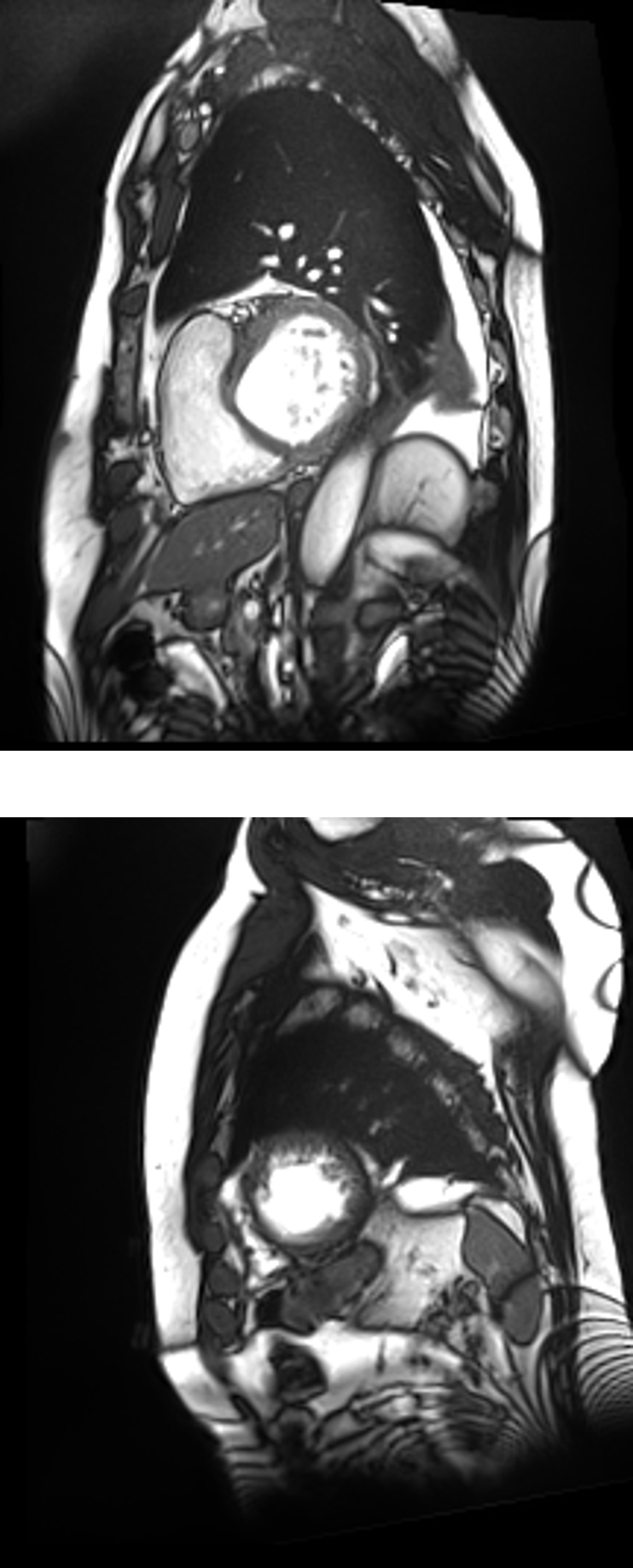}
  }
  \subfloat[]
  {
      \includegraphics[width=0.36\linewidth]{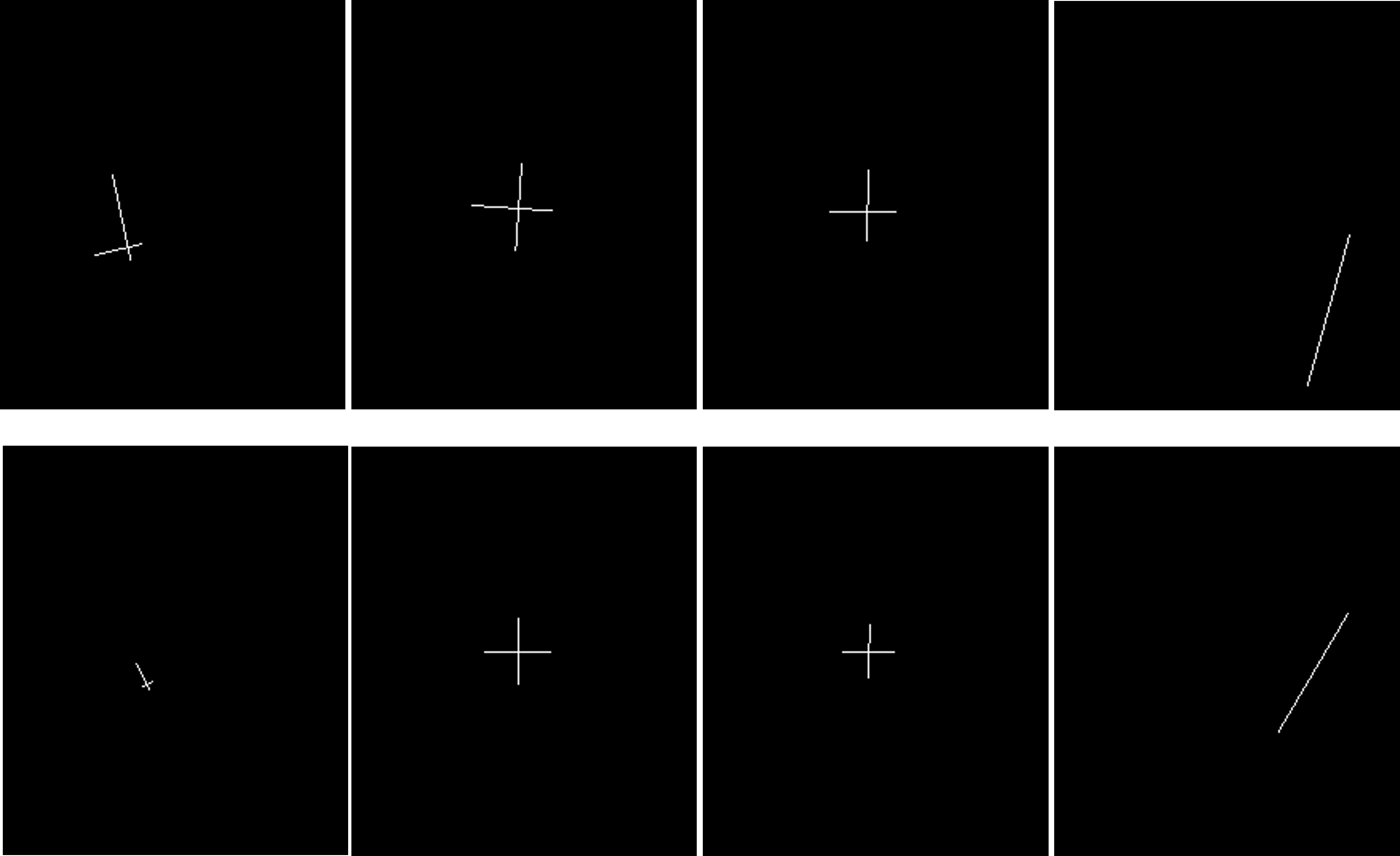}
  }
    \subfloat[]
  {
      \includegraphics[width=0.27\linewidth]{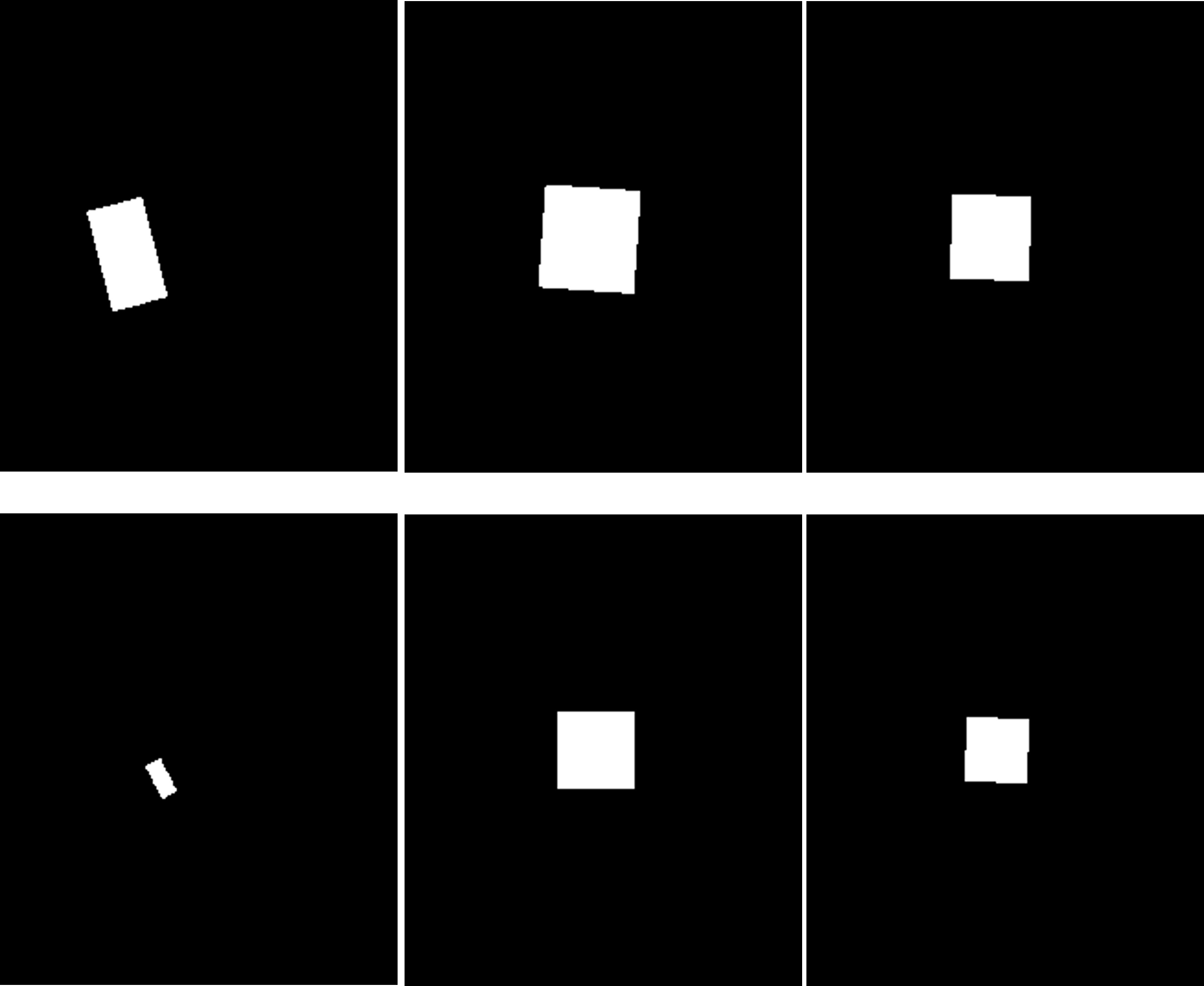}
  }
    \subfloat[]
  {
      \includegraphics[width=0.18\linewidth]{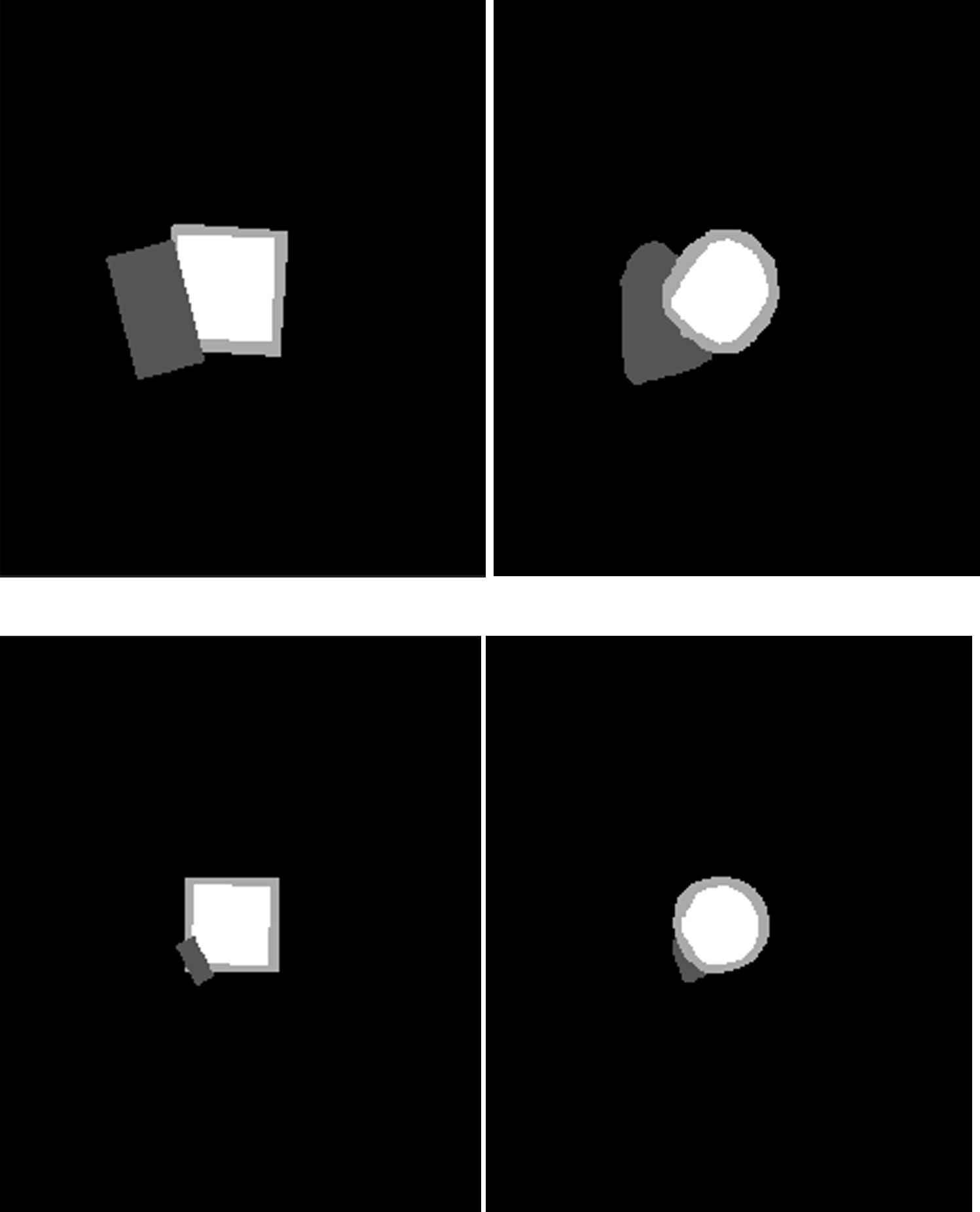}
  }
  \caption{Two examples from the ACDC training set featuring cross-shape scribble annotations and corresponding pseudo masks. 
  (a) Input image 
  (b) From left to right are cross-shape scribbles for Cat 1, 2, and 3, and the segment for Cat 0 (background area). 
  (c) From left to right are the generated pseudo masks based on cross-shape scribbles for Cat 1, 2, and 3, respectively. 
  (d) Left: The combined pseudo masks of all categories. Right: The groundtruth full mask. Dark grey, light grey, and white denote Cat 1, 2 and 3, respectively.}
  \label{fig: supp acdc scribbles}     
\end{figure*}

We use the same tool and follow the same strategy to annotate each category within the ACDC dataset.
It is worth noting that, given that the part of Cat 3 resides entirely within Cat 2, the length of the crossing segments of Cat 2 should be longer than those of Cat 3 to ensure it has sufficient groundtruth information; otherwise, the pseudo mask of Cat 2 can be obscured by the part of Cat 3.
In essence, the cross-shape annotation of Cat 2 not only encompasses the positive pixels but also accounts for the surrounding 'hollow' areas in a ring-like fashion, preserving essential ground truth information for subsequent pseudo mask generation.
\Cref{fig: supp acdc scribbles} (b) presents the cross-shape scribble annotations of two samples from the ACDC training set.
\subsection{Generation of the pseudo masks}
\begin{figure*}[htbp]
  \centering
  \subfloat[]   
  {
      \includegraphics[width=0.25\linewidth]{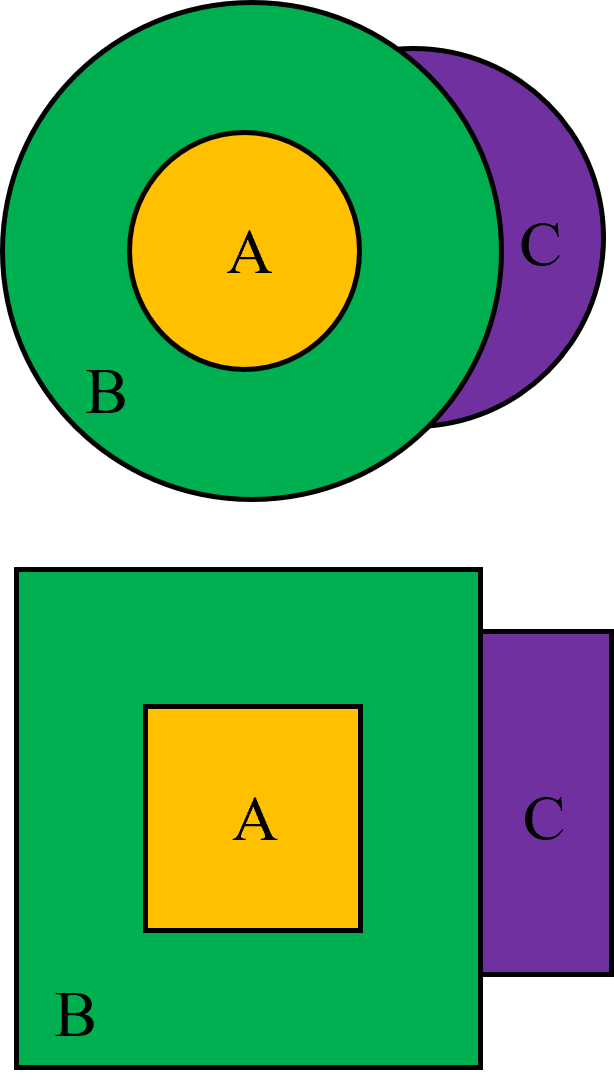}
  }
  \subfloat[]
  {
      \includegraphics[width=0.25\linewidth]{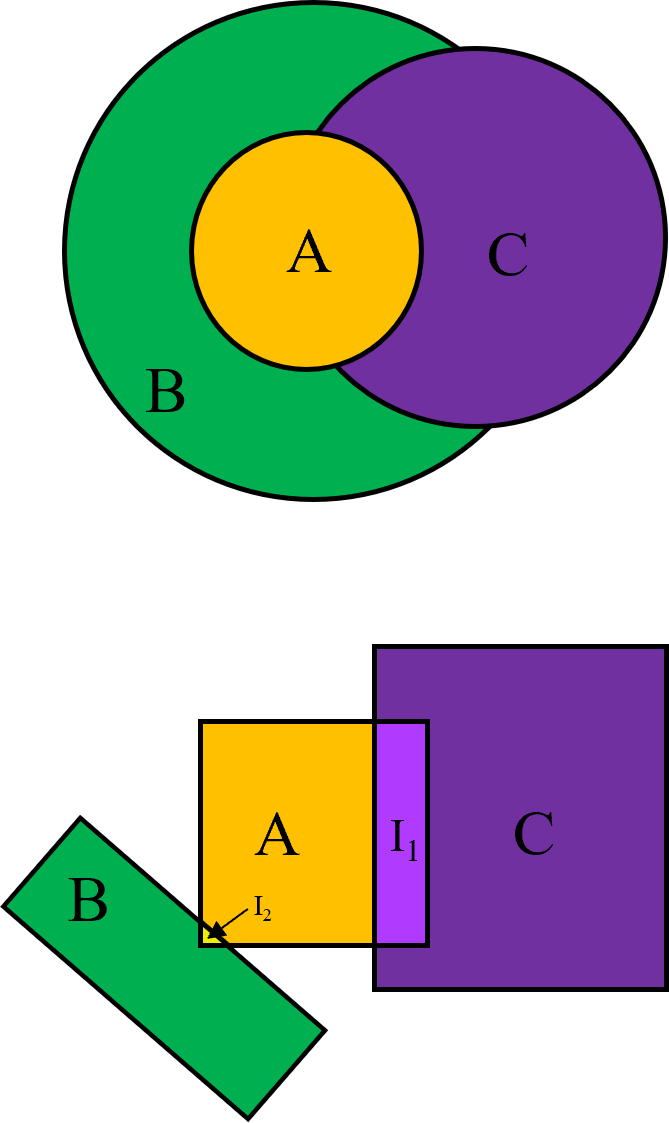}
  }
    \subfloat[]
  {
      \includegraphics[width=0.25\linewidth]{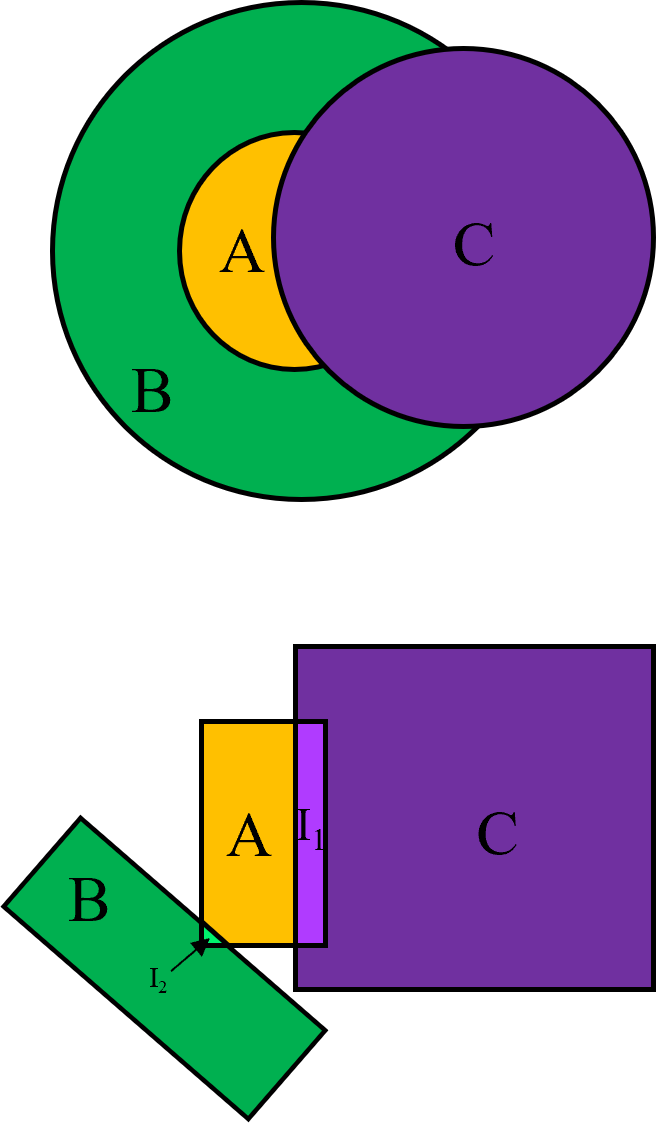}
  }
  \caption{The sketch of three types of inter-class occlusions.
  $A, B$ and $C$ denote the parts of three categories, which are yellow, green and purple.
  $I_1$ and $I_2$ are the overlapping areas of $A, C$ and $A, B$ respectively.
  The first row shows the three patterns of occlusions between the three categories.
  The second row shows the combined pseudo masks generated according to cross-shape scribbles.
  (a) Part $A$ is entirely enclosed within the ring-like part $B$. 
  (b) No ring-like shape if formed. The individual pseudo mask for the yellow category is $A\cup I_1 \cup I_2$. The overlapping areas $I_1$ and $I_2$ are assigned to the purple and yellow categories, respectively.
  (c) The overlapping areas $I_1$ and $I_2$ are assigned to the purple and green categories, respectively.
  }
  \label{fig: supp mul cat rules}     
\end{figure*}

Creating category-specific pseudo masks for the ACDC dataset aligns with the previously described strategy.
The only difference is that an additional combined pseudo mask needs to be generated for the case of multiple categories.
This is because, during pseudo mask generation, there may be instances of inter-class overlap. 
To distinguish which category a shared area belongs to, an additional step is introduced to generate a unified combined pseudo mask, ensuring the absence of inter-class overlap.
To determine the allocation of overlapping areas to the respective categories, we employ two rules. These rules serve to maintain clarity and prevent ambiguity.
First, when one category is entirely encompassed within another, forming a ring-like shape, the overlapping area is assigned to the inner category as illustrated in \Cref{fig: supp mul cat rules} (a).
Second, if two categories share the area, it will be assigned to the category with the largest size.
For example, let $S_p, S_y, S_g$ be the size of the purple, yellow and green categories, respectively.
In \Cref{fig: supp mul cat rules} (b), we have the relationship $S_p>S_y>S_g$, so the shared area I$_1$ is assigned to the purple category and I$_2$ is assigned to the yellow category. 
Consequently, the final pseudo masks for the green, purple and yellow categories become $B$, $C\cup I_1$ and $A\cup I_2$, respectively.
In \Cref{fig: supp mul cat rules} (b), the relation becomes $S_p>S_g>S_y$, so now I$_2$ is assigned to the green one.
Thus, the final pseudo masks for the green and yellow categories are $B\cup I_2$ and $A$.
The second rule ensures the overall accuracy of each category.
If the shared area is mistakenly assigned to a category with the largest size, this category still has sufficient information to correct learning.
However, if it is mistakenly assigned to a small category, the number of correctly labelled pixels may be less than that of the wrong ones.
Consequently, this misallocated information has the potential to disrupt the learning process by introducing incorrect information that is harder to rectify.

\subsection{Network structure}
\label{sec: supp network structure}
\begin{figure*}[htb]
    \centering
    \includegraphics[width=\linewidth]{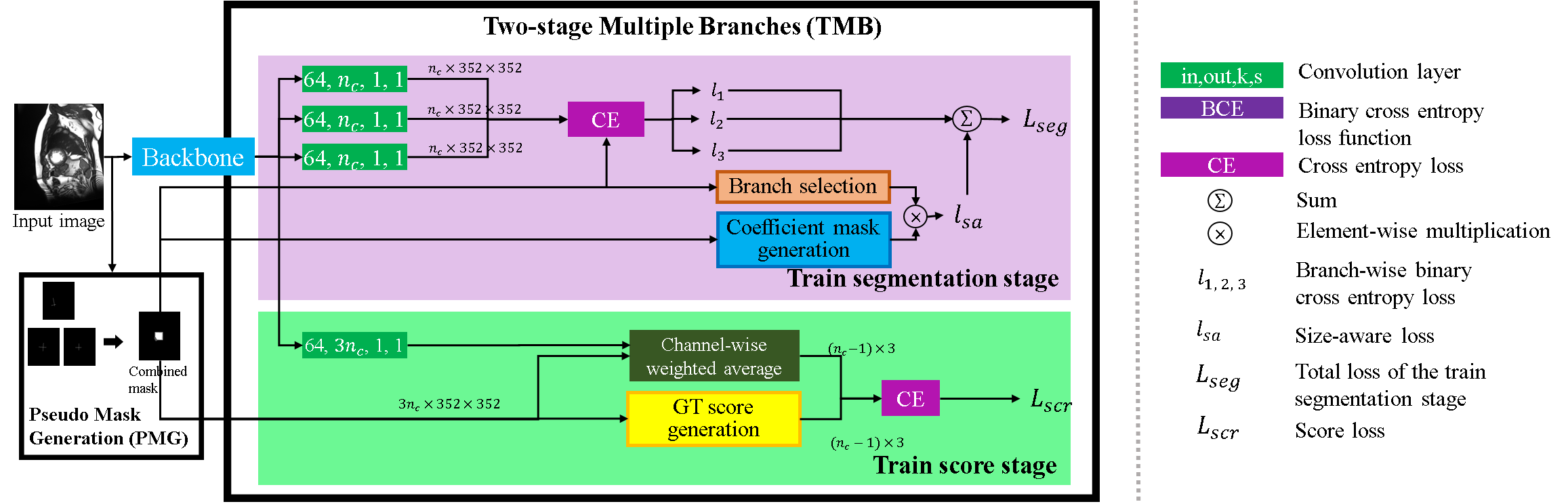}
    \caption{
   The architecture of the proposed size-aware multi-branch method.
   The network consists of the Pseudo Mask Generation (PMG) module and the Two-stage Multiple Branches (TMB) module.
   The TMB module comprises the train segmentation stage and the train score stage.
   The detailed structure of the Coefficient Mask Generation (CMG), Branch Selection (BS), GT Score Generation (GSG) and the Channel-wise Weighted Average (CWA) blocks are presented in \Cref{sec: supp network structure}.
   The number of categories, including the background, is denoted as '$n_c$'.
   For the ACDC dataset, $n_c=4$.
   Note that the shapes of the intermediate outputs are based on the batch size of one.
   }
    \label{fig: supp network mul-cls}
\end{figure*}

\Cref{fig: supp network mul-cls} provides an overview of the network structure for both the pseudo-mask and size-aware multi-branch approaches.
While the overall framework shares similarities with the binary category case, there are notable distinctions, including modifications to the four convolution layers following the backbone, variations in the choice of loss functions, and specific adjustments within the Coefficient Mask Generation (CMG), Branch Selection (BS), GT Score Generation (GSG), and Channel-wise Weighted Average (CWA) blocks.
Consider the size-aware multi-branch approach and take an image from the ACDC training set as an example.
Consider an input grey-scale image $I$ with dimensions $H\times W$.
In the case of an RGB image, the number of the input channels of the backbone should be set to 3.
The input image $I$ is fed into the UNet, which serves as the backbone.
Note that the final single convolution layer in UNet, typically used for pixel-level classification is replaced by three parallel convolution layers with the parameters $c_i=64, c_o=4, k=1, s=1$, along with an additional convolution layer with $c_o=12$, where $c_i, c_o$ are the numbers of input and output channels, $k$ is the kernel size and $s$ is the stride length.
The parallel branches produce the segmentation masks $p_1, p_2, p_3$ with the shape of $4\times H \times W$.
Each predicted mask is responsible for the segmentation of the small, medium or large target.
Then, the cross-entropy losses of each branch, $l_1, l_2, l_3$, are calculated as $l_i=CE(p_i, m_c), i=1,2,3$ where $m_c$ stands for the combined mask.
The operations within the current branch selection block and the coefficient mask generation block are similar to the case of the binary category scenario.
The only difference is that the relative size $r_s$ and the coefficient mask for each category are calculated separately.
Thus, each category may select a different branch to form the final output $p_o$ with dimensions of $4\times H \times W$.
After that, the final output is multiplied by the coefficient mask and averaged to produce the size aware loss $l_{sa}$.
For the train score stage, only the parameters in the additional convolution layer can be adjusted, as all other layers remain fixed.
This layer produces the score mask $s_m$ with dimensions $12\times H \times W$, where the number '12' results from multiplying the number of branches (3) by the number of categories (4).
This implies that within $s_m$, each channel is dedicated to predicting the likelihood of each category selecting the outcome produced by each branch as the optimal match.
Subsequently, both the score mask $s_m$ and the combined mask are forwarded to the GT score generation and channel-wise weighted average blocks.
The procedures within these two blocks closely align with those outlined in the main content.
However, it's important to note that the generated ground truth score takes the form of a matrix with dimensions $3\times3$ rather than a vector, as does the output of the channel-wise weighted average block.
It's worth mentioning that the score matrix excludes the scores associated with the background area to ensure equitable learning across categories.
Subsequently, the output score matrices are sent to the cross-entropy function to derive the loss, denoted as $L_{scr}$, for the train score stage.

\subsection{Experiment settings and results}
The parameter settings for our experiments are as follows.
The values of $\sigma_{i,j}$ are set to infinity and the maximum coefficient $coe$ is set to 10, consistent with the choices made for polyp datasets.
Three branches are used with threshold pairs for four categories as presented in \Cref{fig: supp thresholds acdc}.
The batch size is 4, and 15\% of the training images are used to form the validation set.
The same input size, data augmentation techniques, optimizer and GPU devices are used for training and inference.
For the pseudo mask approach, the learning rate is set at 3e-4.
In contrast, the size-aware multi-branch approach follows a two-stage training process: firstly, the network undergoes 23 epochs of training in the train segmentation stage, with a learning rate of 1e-4.
Subsequently, the model parameters are frozen, and the score map prediction layer is trained with a learning rate of 1e-4 in the train score stage.

\begin{figure*}[htbp]
  \centering
  \subfloat[]   
  {
      \includegraphics[width=0.08\linewidth]{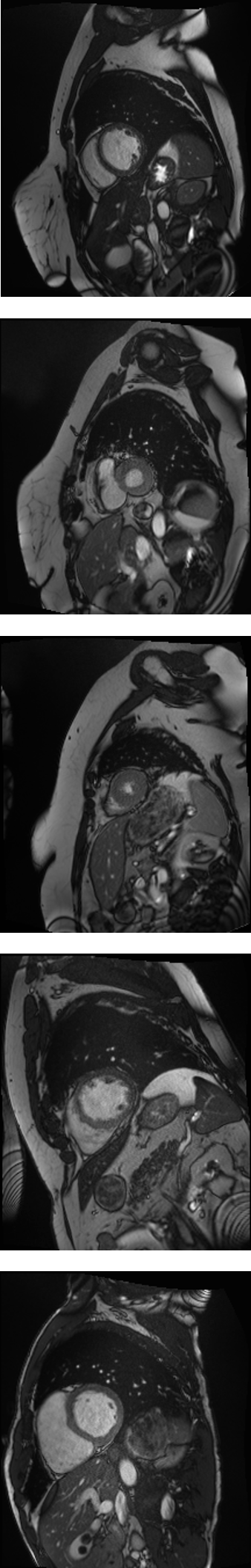}
  }
  \subfloat[]
  {
      \includegraphics[width=0.08\linewidth]{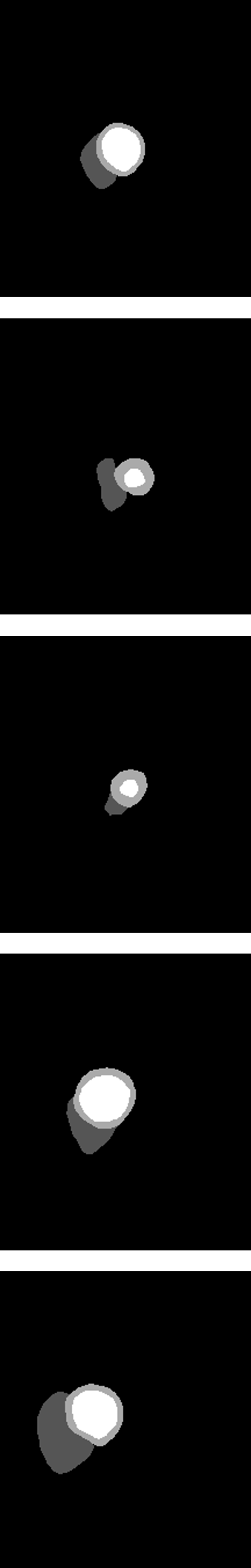}
  }
    \subfloat[]
  {
      \includegraphics[width=0.25\linewidth]{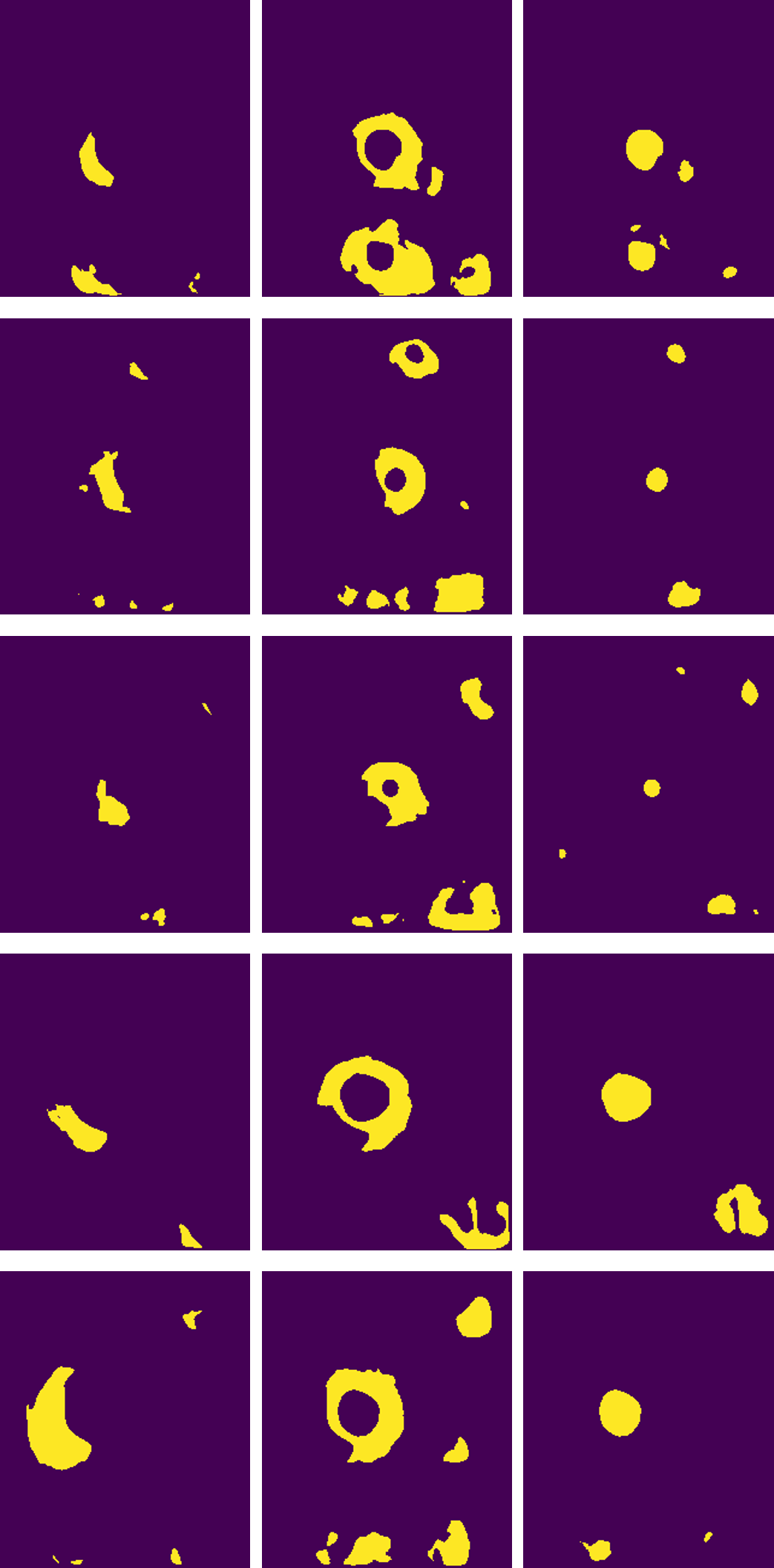}
  }
    \subfloat[]
  {
      \includegraphics[width=0.25\linewidth]{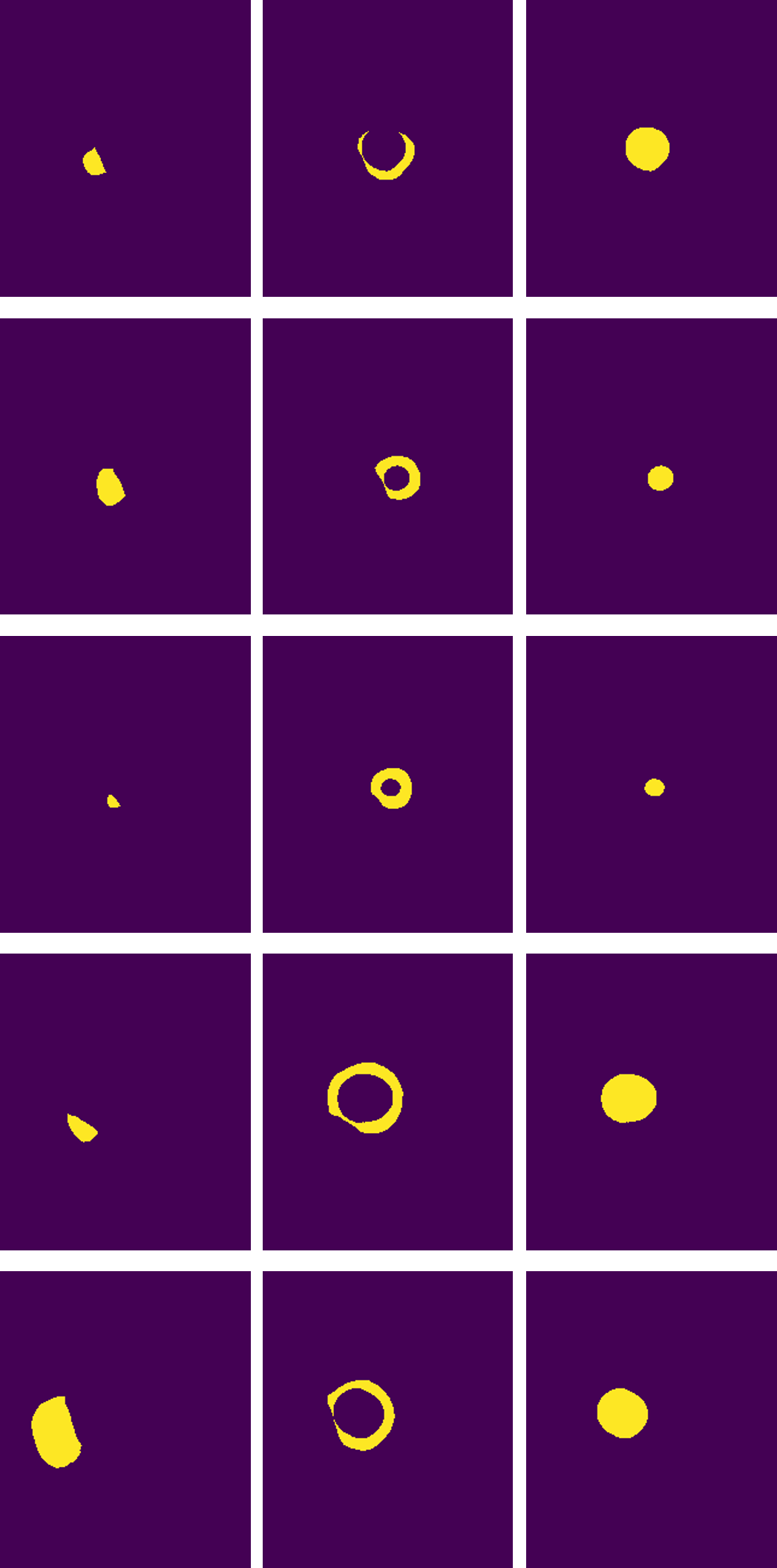}
  }
      \subfloat[]
  {
      \includegraphics[width=0.25\linewidth]{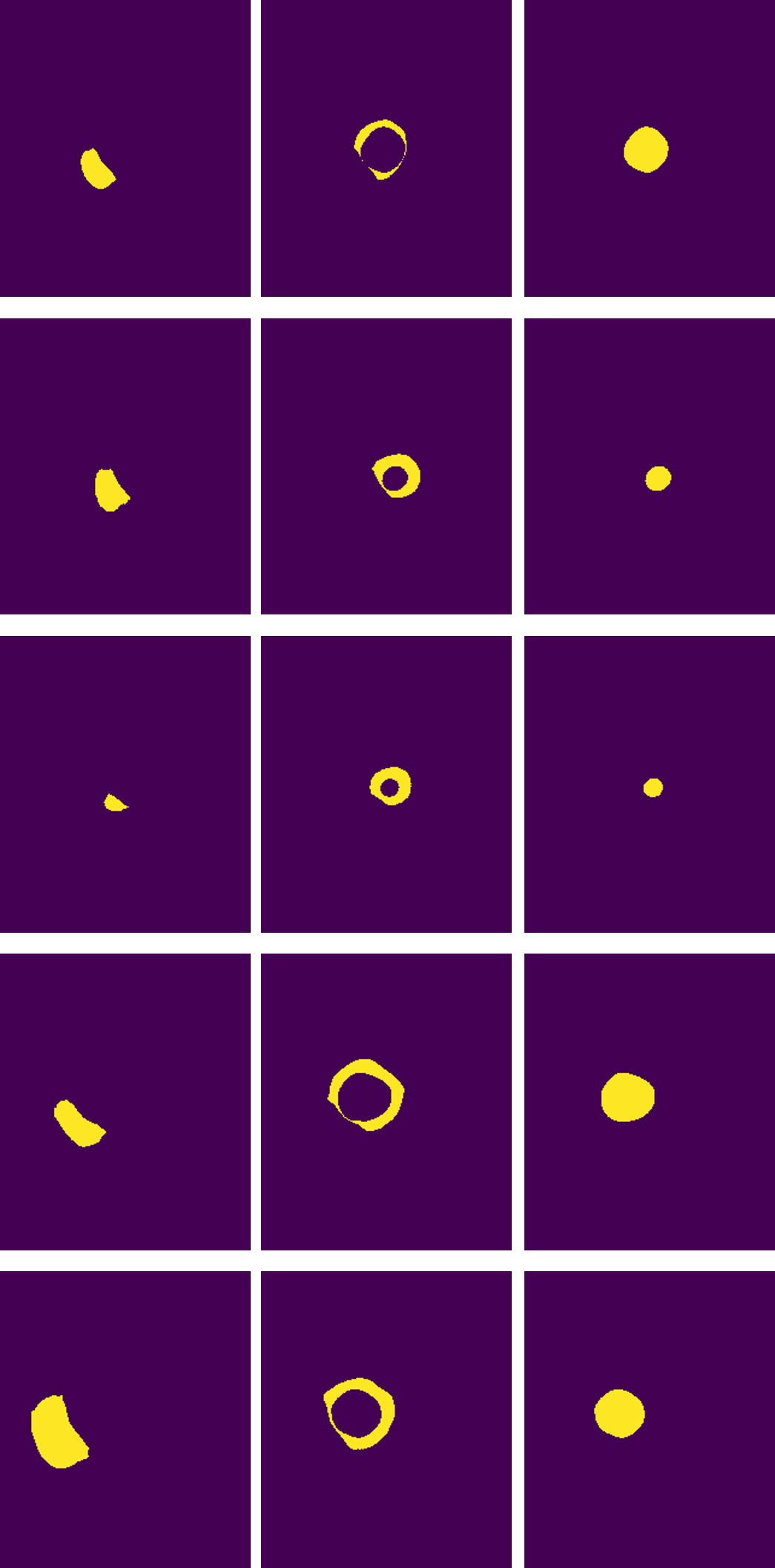}
  }
  \caption{Examples of outputs of different methods on the ACDC test set.
  (a) The input test images.
  (b) Groundtruth. The three categories, Cat 1, 2 and 3, are in dark grey, light grey and white, respectively.
  (c) Segmentation masks predicted by UNet. From left to right are the results for Cat 1, 2, and 3 if not otherwise specified.
  (d) Segmentation masks predicted by UNet combined with the sole PMG module.
  (e) Segmentation masks predicted by UNet combined with the PMG and TMB modules.}
  \label{fig: supp acdc samples}     
\end{figure*}

In, \Cref{fig: supp acdc samples}, we provide visual comparisons of results obtained using the baseline UNet and our proposed approaches on the ACDC dataset.
The proposed methods significantly enhance the accuracy across all categories by reducing the occurrence of false positive pixels.
The inclusion of pseudo masks contributes to masks that closely resemble ground truth, with the size-aware multi-branch method demonstrating particularly significant improvements, especially for Cat 1.

\begin{table}[htb]
\centering
\caption{The thresholds for the four categories of the ACDC dataset utilized in the branch selection block.}
\begin{tabular}{@{}ccccc@{}}
\toprule
      & Cat 0   & Cat 1   & Cat 2   & Cat 3   \\ \midrule
$thr_1$ & 0.94286 & 0.00946 & 0.0155  & 0.00958 \\
$thr_2$ & 0.9657  & 0.01966 & 0.02316 & 0.01804 \\ \bottomrule
\end{tabular}
\label{fig: supp thresholds acdc}
\end{table}

\clearpage
\bibliographystyle{IEEEtran}
\bibliography{main}

\end{document}